\documentclass{article}

\usepackage{authblk}
\usepackage{tabularx,booktabs}
\usepackage{fullpage}
\usepackage{graphicx}
\usepackage{lipsum}
\usepackage[utf8]{inputenc}
\usepackage[margin=1.0in]{geometry}
\usepackage[hidelinks]{hyperref}
\usepackage[T1]{fontenc}
\usepackage{amsmath,amssymb}
\usepackage{bm}
\usepackage{algorithm}
\usepackage{algorithmic}
\usepackage{placeins}
\usepackage{dsfont}
\usepackage{url}
\usepackage{svg}
\usepackage[export]{adjustbox}
\usepackage{todonotes}
\usepackage{soul}
\usepackage{bbding}
\usepackage[labelfont=bf]{caption}
\usepackage{cleveref}
\usepackage{multirow}
\usepackage{comment}
\usepackage{siunitx}
\usepackage{enumitem}
\usepackage{wrapfig}
\usepackage{grffile}
\usepackage[symbol]{footmisc}
\renewcommand{\thefootnote}{\fnsymbol{footnote}}

\begin{document}
\thispagestyle{empty}

\title{Can We Challenge Open-Vocabulary Object Detectors with Generated Content in Street Scenes?}

\setcounter{Maxaffil}{2}
\author[1]{Annika Mütze \textsuperscript{*}}
\author[1, 2]{Sadia Ilyas \textsuperscript{*}}
\author[1]{Christian Dörpelkus}
\author[1]{Matthias Rottmann}

\affil[1]{\small University of Wuppertal, Germany}
\affil[2]{\small Aptiv Services Deutschland GmbH, Wuppertal}
\begingroup
\renewcommand\thefootnote{*}
\footnotetext{Equal contribution. Corresponding author email: \href{mailto:muetze@uni-wuppertal.de}{muetze@uni-wuppertal.de}}
\endgroup
\date{}

\maketitle

\begin{abstract}
Open-vocabulary object detectors such as Grounding DINO are trained on vast and diverse data, achieving remarkable performance on challenging datasets. Due to that, it is unclear where to find their limitations, which is of major concern when using in safety-critical applications. Real-world data does not provide sufficient
control, required for a rigorous evaluation of model generalization.
In contrast, synthetically generated data allows to systematically explore the boundaries of model competence/generalization.
In this work, we address two research questions: 1) Can we challenge open-vocabulary object detectors with generated image content? 2) Can we find systematic failure modes of those models? 
To address these questions, we design two automated pipelines using stable diffusion to inpaint unusual objects with high diversity in semantics, by sampling multiple substantives from WordNet and ChatGPT. On the synthetically generated data, we evaluate and compare multiple open-vocabulary object detectors 
as well as 
a classical object detector. The synthetic data is derived from two real-world datasets, namely LostAndFound, a challenging out-of-distribution (OOD) detection benchmark, and the NuImages dataset. 
Our results indicate that inpainting can challenge open-vocabulary object detectors in terms of overlooking objects. 
Additionally, we find a strong dependence of open-vocabulary models on object location, rather than on object semantics. This provides
a systematic approach to challenge open-vocabulary models and gives valuable insights on how data could be acquired to effectively improve these models.
\end{abstract}

\section{Introduction}
\label{sec:intro}
\begin{figure}[tb]

    \centering 
    \setlength{\tabcolsep}{2pt}
    \begin{tabular}{cccc}
        \includegraphics[width=0.23\linewidth]{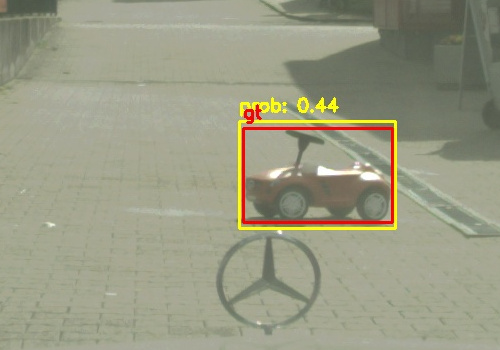}%
        &
        \includegraphics[width=0.23\linewidth]{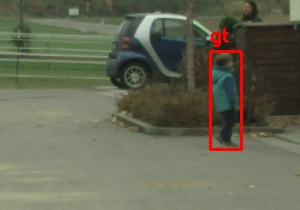}%
        &       

        \includegraphics[width=0.23\linewidth]{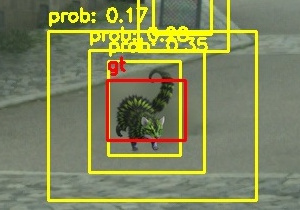}%
        &
        \includegraphics[width=0.23\linewidth]{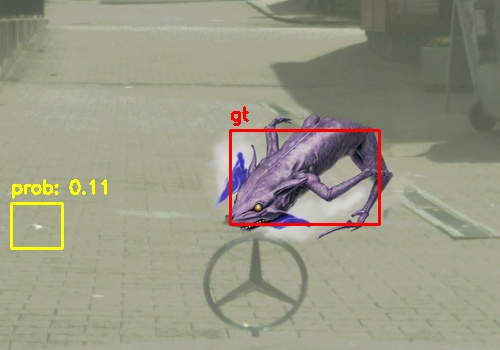}%
        \\
        real & real & inpainted & inpainted%
    \end{tabular}

    \caption{Range of detection capability of Grounding DINO on LostAndFound. All panels are prompted with ``object on the street'' and detections are visualized in yellow. From left to right: 1) detected real object; 2) overlooked real object; 3) detected synthetic object; 4) overlooked synthetic object where, however, small leaves by side are detected.}
    \label{fig:page1_predictions}
\end{figure}

    Open-vocabulary models have recently become popular due to their strong generalization capabilities and zero-shot performance. Open-vocabulary object detectors, like Grounding DINO \cite{liu2023grounding} and YOLO-World \cite{cheng2024yolo} mostly outperform classical object detectors \cite{kamath2021mdetr, zhang2022glipv2, zhong2022regionclip} on common benchmark datasets, such as COCO-O \cite{mao2023coco} and Cityscapes \cite{Zhang_2023_ICCV}. Classical object detectors tend to fail when evaluated on data not closely aligned with their training data distribution. On the other hand, open-vocabulary object detectors have been applied with greater success to out-of-distribution (OOD) scenarios, showing consistently better results across various datasets \cite{ilyas2024potential}. Considering the large and diverse amount of data used to train these models, it is unclear where to find the limitations of their generalization capabilities. In particular, it is also unclear how to acquire the data that challenge these models.
A possible approach is to use uncertainty estimation methods in an online fashion during data acquisition \cite{miller2018dropout,miller2019merging,schubert2021metadetect,riedlinger2023gradient} to filter data where a given model is uncertain, 
or detecting semantically unusual objects or anomalies \cite{du2022vos,wilson2023safe,ilyas2024potential}.
In this work, we follow a different approach, which can be used offline without prior extensive data acquisition. With the advent of generative diffusion models, such as text-to-image models with inpainting capabilities, it is possible to generate high quality image content \cite{saharia2022photorealistic, ramesh2022hierarchical, nguyen2024dataset}. This raises the question \emph{whether we can challenge open-vocabulary object detectors with generated image content.}
The answer may not be straightforward
as it is also a question of generalization capabilities, and, both open-vocabulary object detectors and diffusion models are trained on an enormous amount of data. 
To study our research question, we construct two distinct carefully designed generic procedures to automatically inpaint unusual objects or concepts into street scenes. Therein, for the first procedure we sample multiple substantives from WordNet \cite{miller1995wordnet} to create versatile object semantics, while the second procedure relies on ChatGPT to generate objects.
We evaluate open-vocabulary object detectors; \cite{liu2023grounding,cheng2024yolo, zhao2024real, kamath2021mdetr} and a classical object detector \cite{ren2016fasterrcnnrealtimeobject} on our synthetically generated datasets including LostAndFound \cite{pinggeralost}, a dataset originally designed for anomaly detection and NuImages, a dataset for autonomous driving \cite{caesar2020nuscenes}.
Our proposed setup offers a framework that enables the systematic evaluation of these models by varying the data.
The results obtained show that synthetically generated data can induce false negatives in open-vocabulary object detectors nearly as often as observed with real-world data, shown in \cref{fig:page1_predictions}. 

We identify challenging street scene configurations for open-vocabulary object detectors and observe clear influences of both prompt formulation and object placement in the images.
Interestingly, the location of an inpainted object and the prompts used appear to have a greater impact on the model's performance than the semantic content of the generated objects.

Our contributions are summarized as follows:
\begin{itemize}

    \item We propose two generic inpainting procedures to generate synthetic image data and create a test bench for state-of-the-art open-vocabulary models in street scenes.
    \item We apply our generic inpainting procedures to LostAndFound and NuImages, testing a number of open-vocabulary object detectors on a variety of unusual image content with rich semantic diversity.
    \item We observe that the choice of scene and location of an inpainted object influence the detection performance. Closer inspection reveals that the location of the inpainted object is a highly influential factor while the object’s semantic class is less relevant.
\end{itemize}
   
\section{Related Work}
  This section gives a brief overview of work related to synthetic data generation methods, synthetic data in image recognition and development in open-vocabulary object detection.

\textbf{Synthetic Data using Latent Diffusion.}
The performance of image generation models experienced an impressive improvement due to text-to-image diffusion models such as GLIDE \cite{nichol2021glide}, Stable Diffusion (SD) \cite{rombach2022high} and DALL-E 3\cite{betker2023improving}. 
Recent advancements in diffusion models have enabled high quality image synthesis and inpainting \cite{xie2023smartbrush,lugmayr2022repaint, corneanu2024latentpaint}, with Stable Diffusion being a widely used framework. IP-Adapter \cite{ye2023ip-adapter} combines image and text prompts to achieve multimodal image synthesis. 
Interfaces like AUTOMATIC1111 \cite{AUTOMATIC1111_Stable_Diffusion_Web_2022} allow easy access to both text-to-image and image-to-image tasks. ImagenHub benchmark \cite{ku2023imagenhub} evaluates Stable Diffusion models across various image generation tasks, concluding superior performance in text-to-image generation tasks.

\textbf{Synthetic Data in Image Recognition.}
Synthetic data is particularly useful in the field of image recognition, specially when annotated real-world data is hard to obtain. It provides a beneficial alternative for many machine learning applications. Previous work suggests that including synthetic data within the training process can boost the detection performance on different benchmarks \cite{liu2023utilizing, gui2024remote, nowruzi2019realdataactuallyneed, lin2023explore, 12132_2022}. Therefore, the diversity of image examples seems more important than being photo-realistic. 
The method proposed in \cite{liu2024can} generates synthetic out-of-distribution (OOD) images by inpainting, using prompts that are similar to the original in-distribution objects. By incorporating these synthetic images into the training set, models such as VOS \cite{du2022vos} and Faster R-CNN \cite{ren2016fasterrcnnrealtimeobject} achieved an increase in performance.
Benchmark like Fishyscapes \cite{blum2021fishyscapes} designed for anomaly segmentation, has shown to utilize synthetic data to test models under domain shifts. Another work \cite{loiseau2024reliability} investigates the reliability of synthetic data for estimating the robustness of pre-trained segmentation models to edge cases in the real-world, finding that the performance of models when evaluated on their synthetic OOD data highly correlates to the performance on real OOD data. Another study demonstrates strong correlation for cars and pedestrians by assessing the transferability of testing results from synthetic to real-world data \cite{rosenzweig2021validation}.
In contrast to previous works, we deliberately create systematically abnormal objects to challenge particularly open-vocabulary object detectors. While many of the previous works focus on training models on synthetic data to improve the performance, our approach evaluates these models on synthetic data to identify the content on which open-vocabulary object detectors fail.
Recent open-vocabulary models such as as GLIP \cite{li2022grounded}, OWL-ViT \cite{minderer2022simple}, Grounding DINO \cite{liu2023grounding}, MDETR \cite{kamath2021mdetr}, YOLO-World \cite{cheng2024yolo}, OmDet \cite{zhao2024real}, have received significant attention due to their robustness in various downstream applications. 
Standard benchmarks for evaluating open-vocabulary detection performance include COCO \cite{lin2014microsoft}, LVIS \cite{gupta2019lvis}, Objects365 \cite{shao2019objects365}. 
A study reveals that these models performing well on standard benchmarks, tend to struggle with fine-grained distinction of object properties \cite{bianchi2024devil}.
As a complementary perspective, we rather evaluate open-vocabulary object detectors in a synthetically generated framework. That enables fine-grained control over the object placement in the scene and allows for creating semantically diverse objects. Making a challenging test-bench to systematically study the model behavior.

\section{Inpainting Procedure}
    
In this work, we are mainly interested in the performance of open-vocabulary object detectors on synthetic data. Detailed information about the models used in our experiments is provided in appendix A. In the following, we describe the inpainting process to generate synthetic datasets in detail. 

\begin{figure}[htbp]
    \centering
    \includegraphics[width=8cm,height=4cm]{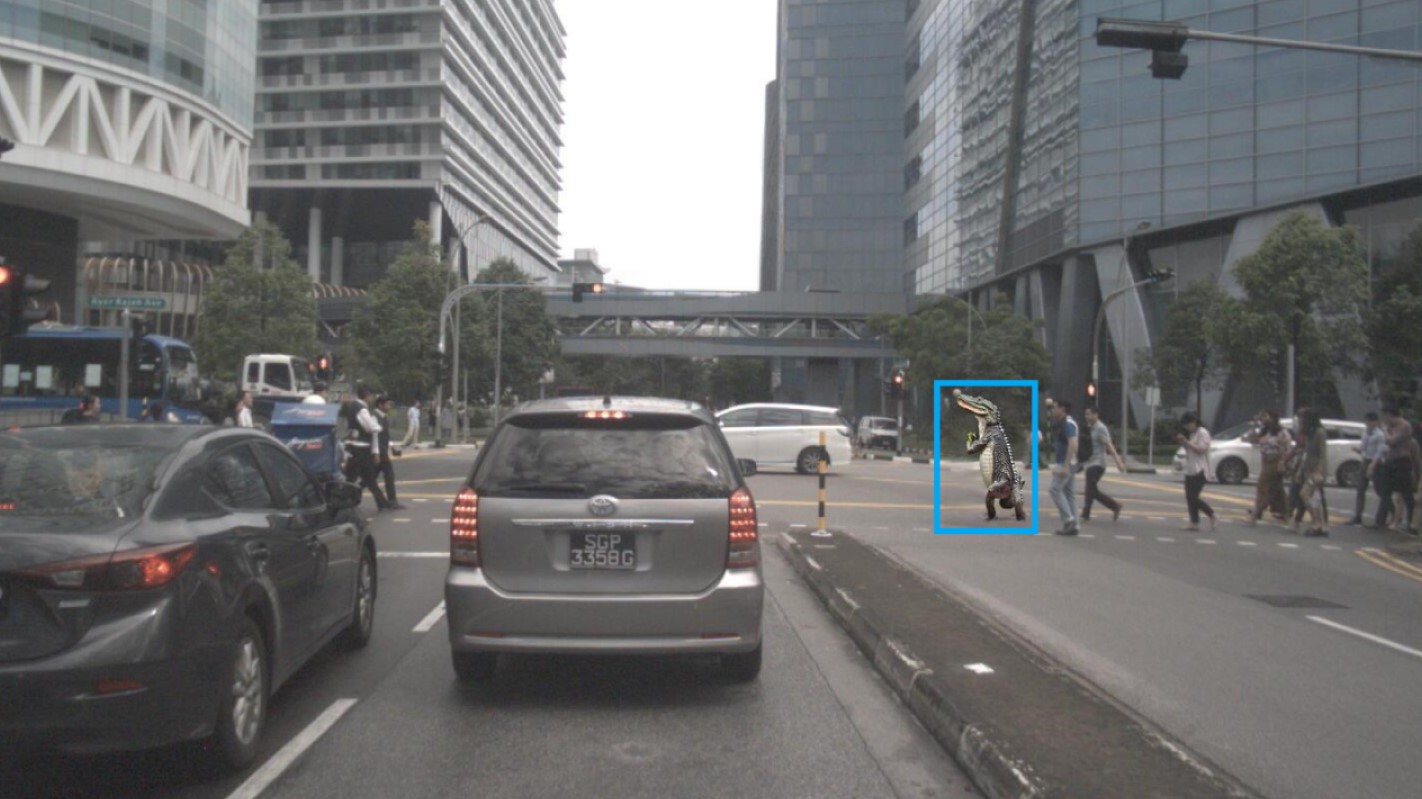}
    \includegraphics[width=8cm,height=4cm]{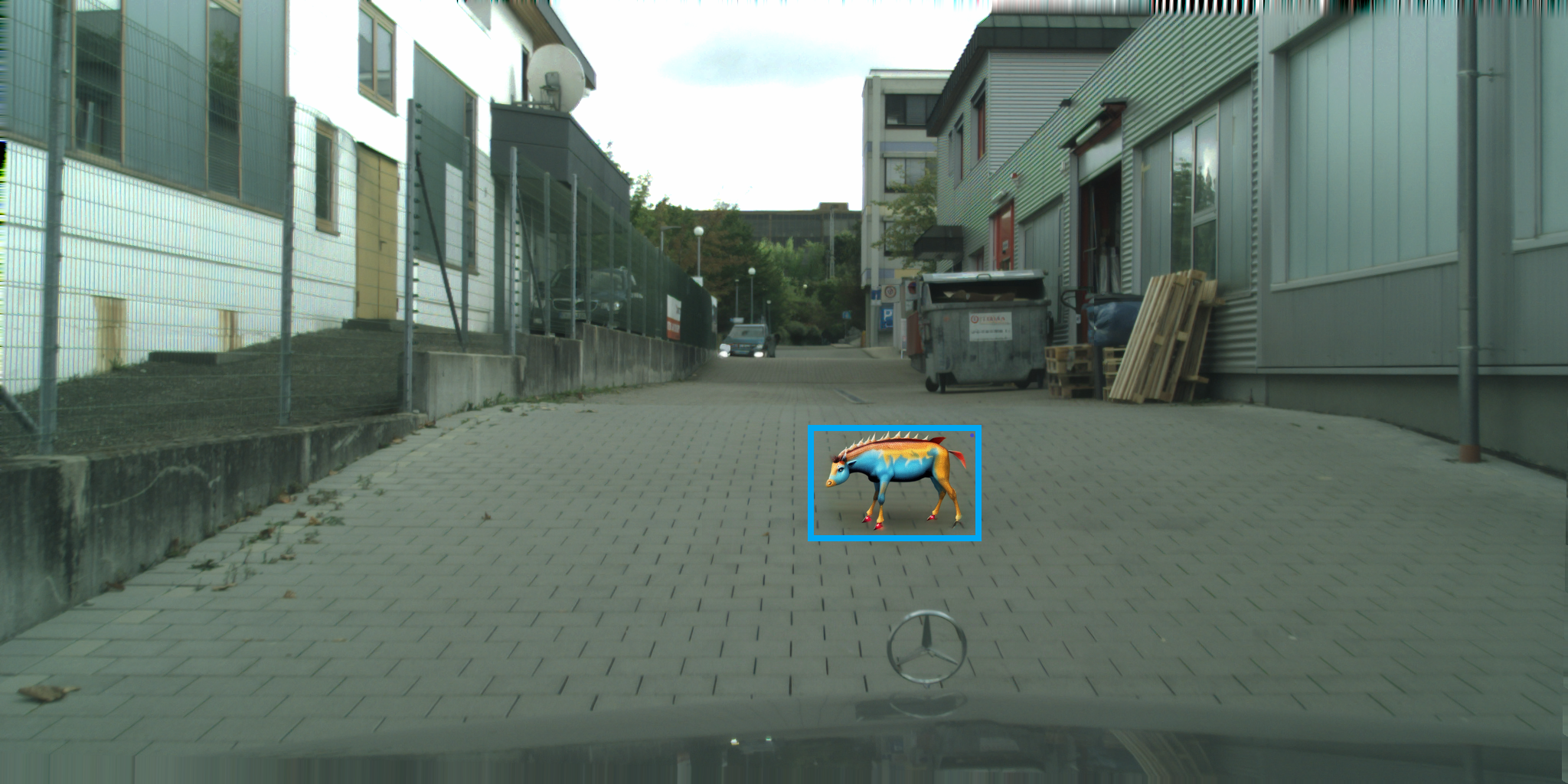}
    \caption{The left image shows the Single-concept inpainting on NuImages and the right one shows the Hybrid-concept inpainting. The blue bounding box indicates the inpainted object.}
    \label{fig:inpainting}
\end{figure}

\paragraph{Hybrid-concept inpainting.} In our first inpainting procedure that is supposed to maximize semantic variety, we proceed as follows: Starting with a dataset, we first filter images that contain exactly one object in the annotated ground truth. Additionally, we only consider images where the object size is at least 3,000 pixels, ensuring that the inpainted object is large enough to be evaluated also by the human eye. To obtain abnormal objects and concepts, we sample up to four random words from WordNet \cite{miller1995wordnet} and combine them to a prompt as follows "word\textsubscript{1}\_word\textsubscript{2}\_word\textsubscript{n}\_hybrid".
Instead of just directly using the ground-truth bounding box (bbox) as a mask to inpaint, we create an oval mask based on the original bbox. 
In our experiments, we observed unrealistically razor-sharp inpaintings. We experimented with passing the whole image or just a $512 \times 512$ frame around the bounding box center to the inpainting models. The second option, by which we proceed, not only led to smoother inpaintings, but additionally, the computing time was significantly lower.
After inpainting, the frames are simply placed back into the original image, \cref{fig:inpainting} shows an example image in this setting.
In our experiments, the inpainting failed in some cases in terms of not producing an actual object, only slightly changing the background or even removing an original object, we manually revised the data and discarded these images from the generated dataset. 

\paragraph{Single-concept inpainting.} For the second procedure, we obtained the object classes using ChatGPT, by prompting it to return ``a list of objects that are unusual for street scenes''.
The lists mainly contained animals and household items, considered as unusual for street scenes. 
In order to inpaint the objects on drivable region, we randomly selected a target object exceeding a minimum overlap requirement with the drivable surface, i.e., the road, and replaced it with a synthetic object.
The minimum size of the object was again at least 3,000 pixels.
To achieve higher quality inpaintings, we noticed that the size of the crop region around the target object to be replaced had a direct effect. Thus, we experimented with and selected three different quadratic crop regions, explained in detail in the later section.
Figure \ref{fig:inpainting} shows Single-concept inpainting in NuImages street scene.

\section{Experimental Setup}
  In the following, we briefly introduce our experimental setup to challenge open-vocabulary object detectors with generated content in street scenes. This includes a description of the prompts provided to the detection models, the datasets used as well as a description of the inpainting process to retrieve our proposed synthetic datasets. Moreover, we elucidate the evaluation methods applied to quantify the performance of open-vocabulary object detectors on both the real benchmark data and our generated datasets.

\paragraph{Prompt variation.} 
Our experimental setup includes multiple prompts that we provided to the open-vocabulary models under test. Prompt variations are introduced to evaluate their impact on model performance. At the same time, this helps to ensure open-vocabulary object detectors can reveal their full potential when comparing them with simple detectors. To empower the open-vocabulary object detectors to generalize and detect all different kinds of objects, we kept the prompts vague. However, mainly focus on the aspect of street scenes. This resulted in the following prompts: p1 = ``\textit{object}'', p2 = ``\textit{object . animal . person}'', p3 = ``\textit{object on the street}'', p4 = `` \textit{obstacle on the street}'' and p5 = ``\textit{something on the street}''.

\paragraph{Dataset description.} We consider two base datasets: LostAndFound and NuImages.

The \textbf{LostAndFound} dataset \cite{pinggeralost} consists of street scenes captured at five distinct locations, showing various road surface textures, and pronounced changes in illumination conditions. To serve the task of anomaly detection, this dataset predominantly shows smaller-sized objects, like animals or toys, placed at different distances. 
The test set of LostAndFound is divided into five subsets, each representing a different location with 371, 248, 304, 137, and 143 images, respectively. Test images can show multiple objects, but not all of them are assigned to an actual ground truth mask in the dataset.
To generate our first synthetic dataset \textbf{synthetic LostAndFound w/ hybrid-concept inpainting}, we applied the inpainting processes explained in the previous section to the test images of the LostAndFound dataset. After filtering for images containing only one object with an area of at least 3,000 pixels in the annotated ground truth, we ended up with 179 images. To acquire an adequate number of samples, we repeated the inpainting 10 times per image using a batch size of 2. Thereby, we ended up with a total number of 3,580 images for LostAndFound w/ hybrid-concept inpainting. 
For inpainting, we used here the AUTOMATIC-1111 framework. In total, we created four synthetic LostAndFound datasets using slightly different parameters when initializing the stable diffusion-based dreamshaper\_8inpainting~\cite{-_2023} inpainting model. We refer to the appendix A for the implementation details.

The \textbf{NuImages Dataset} \cite{caesar2020nuscenes} %
is part of the larger NuScenes dataset, known for autonomous driving that includes multi modal sensor data. NuImages is a subset of it focusing on the 2D image data captured by six cameras, in particular the 2D annotations. 
To create our dataset \textbf{synthetic NuImages w/ single-concept inpainting}, we used only front-camera images from NuImages. For inpainting we used the runwayml stable diffusion inpainting model from huggingface \cite{runwayml2023inpainting}.
The target object was cropped depending on its size, and the crop size was one of these; $512\times512$, $256\times256$ and $128\times128$ in width and height. That means, if an object bounding box has a dimension greater or equal to $256\times256$ the crop size is $512\times512$, similarly for remaining objects larger than $128\times128$ in dimension, the crop size is set to $256\times256$ and for the remaining smaller objects it is $128\times128$. 
However, before inpainting the cropped regions were upscaled to $512 \times 512$ if smaller than 512, to stay consistent with the input dimensions of the stable diffusion model. That is, the input dimensions are expected to be a factor of 8 and preferred a resolution of $512 \times 512$.
We mostly kept the default parameters and only changed the sampling steps to $80$ and padding mask crop to $32$ which helped in reducing the artifacts around the boundaries and resulted in more seamless inpainting. The prompt used for inpainting consists of a keyword taken from ChatGPT with the remaining fixed phrase, i.e., ‘keyword, high resolution, standing on the road’. The list of keywords taken from ChatGPT is provided in the appendix G. The inpainted images were then manually filtered to remove the failed inpaintings and that resulted in an image count of 4,102 images.

\paragraph{Metrics.}
For evaluating the detection performance, we use the well-known object detection metric Area under the Precision-Recall Curve (AUPRC). AUPRC is reported by evaluating Precision-Recall pairs for different confidence score levels.
Additionally, true positives (TP), false positives (FP), and false negatives (FN) are reported for a fixed score threshold of 0.1 to gain insight into the model performances on specific scenes. 
Predictions were considered true positives (TP) if their IoU with the ground truth was greater than $0.5$. 
Before counting TP and FP, we apply Non-Maximum Suppression (NMS) for each prompt separately to eliminate redundant predictions. We also report Average Precision (AP\textsubscript{50..95}) and Average Recall (AR\textsubscript{50..95}) using standard COCO evaluation.

\section{Analysis of models performance and failure modes} 
    To obtain a comprehensive assessment of the performance of open-vocabulary object detectors on synthetic data, we confronted them with our aforementioned synthetic datasets. Visual examples of TP and FN predictions are provided in appendix F.

\paragraph{Models performance across datasets.}

\begin{table}
\caption{Comparison of detection performance for 5 different models on real LostAndFound (179 images) and synthetic LostAndFound w/ hybrid-concept inpainting (3539 images).}
\scriptsize
\label{tab:LF4}
\setlength{\tabcolsep}{4pt}
\resizebox{\textwidth}{!}{
\begin{tabular}{ll
                *{6}{c} 
                *{6}{c}}
\toprule
\multicolumn{2}{c}{} 
& \multicolumn{6}{c}{\textbf{Real LostAndFound}}
& \multicolumn{6}{c}{\textbf{LostAndFound w/ hybrid-concept inpainting}} \\
\cmidrule(lr){3-8} \cmidrule(lr){9-14}
\textbf{Model} & \textbf{Prompt} 
& \textbf{AUPRC} & \textbf{AP} & \textbf{AR} & \textbf{TP} & \textbf{FP} & \textbf{FN}
& \textbf{AUPRC} & \textbf{AP} & \textbf{AR} & \textbf{TP} & \textbf{FP} & \textbf{FN} \\
\midrule
\multirow{5}{*}{\rotatebox[origin=c]{90}{MDETR}} 
 & p1 & 0.2147 & 0.1262 & 0.4335 & 141 & 812 & 38
 & 0.1935 & 0.0506 & 0.1980 & 2205 & 14700 & 1334 \\
 & p2 & 0.3586 & 0.1714 & 0.4291 & 160 & 635 & 19
 & 0.2130 & 0.0455 & 0.1862 & 2234 & 12470 & 1305 \\ 
 & p3 & 0.2206 & 0.1173 & 0.4464 & 161 & 835 & 18
 & 0.1578 & 0.0402 & 0.2073 & 2457 & 15574 & 1082 \\ 
 & p4 & 0.1657 & 0.1139 & 0.3453 & 104 & 582 & 75
 & 0.0845 & 0.0282 & 0.1859 & 2122 & 13163 & 1417 \\
 & p5 & 0.1230 & 0.0665 & 0.3229 & 120 & 428 & 59
 & 0.0909 & 0.0270 & 0.1693 & 2002 & 8631 & 1537 \\
\midrule

\multirow{5}{*}{\rotatebox[origin=c]{90}{OmDet}}
 & p1 & 0.3658 & 0.2945 & 0.7101 & 152 & 368 & 27
 & 0.3868 & 0.1221 & 0.3012 & 2660 & 8043 & 879\\
 & p2 & 0.1866 & 0.1617 & 0.7860 & 164 & 1486 & 15
 & 0.1034 & 0.0390 & 0.3226 & 2833 & 38376 & 706\\
 & p3 & 0.2726 & 0.1947 & 0.4978 & 126 & 393 & 53
 & 0.3706 & 0.0949 & 0.2503 & 2608 & 7625 & 931\\
 & p4 & 0.1537 & 0.0733 & 0.4341 & 154 & 794 & 25
 & 0.1040 & 0.0246 & 0.1827 & 2253 & 16471 & 1286\\
 & p5 & 0.0209 & 0.0225 & 0.1592 & 39 & 214 & 140
 & 0.0139 & 0.0058 & 0.0595 & 654 & 4271 & 2885\\
\midrule

\multirow{5}{*}{\rotatebox[origin=c]{90}{G. DINO}}
 & p1 & \textbf{0.6520} & \textbf{0.5488} & \textbf{0.7352}
 & 154 & 1888 & 25
 & \textbf{0.6594} & \textbf{0.1975} & \textbf{0.3319} & 2957 & 34340 & 582 \\
 & p2 & 0.1652 & 0.1687 & 0.8196 & 167 & 2540 & 12
 & 0.1103 & 0.0477 & 0.3284 & 2986 & 48919 & 553 \\
 & p3 & 0.3580 & 0.2285 & 0.5654 & 162 & 2033 & 17 
 & 0.3436 & 0.0793 & 0.2293 & 2605 & 33585 & 934  \\
 & p4 & 0.0712 & 0.0465 & 0.4034 & 118 & 1562 & 61
 & 0.0775 & 0.0192 & 0.1887 & 2183 & 30220 & 1356  \\
 & p5 & 0.0518 & 0.0264 & 0.2631 & 100 & 2400 & 79
 & 0.0498 & 0.0123 & 0.1251 & 1618 & 43207 & 1921 \\
\midrule
\multirow{5}{*}{\rotatebox[origin=c]{90}{\shortstack{YOLO\\World}}}
 & p1 & 0.1173 & 0.1213 & 0.1173 & 22 & 0 & 157
 & 0.1014 & 0.0355 & 0.0448 & 381 & 32 & 3158 \\
 & p2 & 0.0765 & 0.0753 & 0.4240 & 84 & 561 & 95
 & 0.0248 & 0.0132 & 0.1128 & 1000 & 11024 & 2539 \\
 & p3 & 0.5866 & 0.4914 & 0.6531 & 133 & 197 & 46
 & 0.2852 & 0.0899 & 0.1984 & 1792 & 3923 & 1747 \\
 & p4 & 0.0434 & 0.0548 & 0.0492 & 9 & 1 & 170
 & 0.0098 & 0.0082 & 0.0050 & 43 & 25 & 3496 \\
 & p5 & 0.0664 & 0.0749 & 0.2101 & 40 & 71 & 139
 & 0.0207 & 0.0116 & 0.0381 & 326 & 1382 & 3213 \\
\midrule

\multirow{1}{*}{FRCNN}
 & - & 0.0241 & 0.0213 & 0.1659 & 37 & 680 & 142
 & 0.0404 & 0.0149 & 0.2529 & 2183 & 40258 & 1356 \\           
\bottomrule
\end{tabular}
}
\end{table}
We compare the model performance across two dataset settings: (i) Real LostAndFound and (ii) LostAndFound w/ hybrid-concept inpainting. 
In \cref{tab:LF4}, the results on these datasets are shown. It can be seen that Grounding DINO performs the best on the real LostAndFound dataset with an AUPRC score of 65\%, additionally reaching an AP of 54\% and AR of 73.5\%. In contrast, Faster RCNN as representative for a classical object detector performs poorly compared to all other models with an AUPRC of less than 2.5\%.
This can be attributed to the fact that open-vocabulary models benefit from strong priors acquired from large-scale vision-language data that equip them with the knowledge of typical object patterns. Nevertheless, all models fail to reliably detect all real objects as indicated by the number of FN for the fixed confidence score threshold of 0.1.
The performance in terms of AUPRC in setting (ii) is surprisingly as good as in setting (i), leading to the conclusion that our synthetic data can challenge Grounding DINO to a similar extent. 
However, there is a drop in the metrics AP and AR when going from real to synthetic objects, declining to 19\% and 33\%, respectively.
This is surprising since the inpainted concepts are clearly visible. However, this finding could be attributed to the lack of photo realism.
Comparing the different prompts across models reveals a strong prompt-dependence for all models. In particular, for a given model the detection performance even varies among generic and semantically similar prompts.
In appendix B, we provide additional results on LostAndFound w/ single-concept inpainting, and NuImages w/ single-concept inpainting.

\paragraph{Model performance across scenes.}
Based on our aforementioned results, we broke down our analysis to image level to relate the performance of the detectors. Therefore we generated three more datasets based on LostAndFound by the hybrid-concept inpainting method using slightly different parameters (see appendix A for details). In particular, we inspected the number of FN results per scene. The original dataset (real) contains each scene 
once, whereas for the synthetic datasets (V1 - V4) each scene was used 20 times to inpaint different generated objects. 
Since we use a batch size of 2 for the inpainting, this yields more than 7,100 inpainted concepts of hybrid semantics to test the prediction capability.
\begin{wraptable}{r}{0.5\textwidth}
    \centering
    \scriptsize
    \caption{Rank correlations between scenes of different datasets; scenes ranked by FN counts.}
    \resizebox{0.45\textwidth}{!}{
    \begin{tabular}{llllll}
    \toprule
     & \textbf{real} & \textbf{V1} & \textbf{V2} & \textbf{V3} & \textbf{V4} \\ 
    \midrule
    \textbf{real} & 1.0 & 0.605 & 0.506 & 0.454 & 0.438 \\
    \textbf{V1} &  & 1.0 & 0.947 & 0.924 & 0.896 \\
    \textbf{V2} &  &  & 1.0 & 0.961 & 0.949 \\
    \textbf{V3} &  &  &  & 1.0 & 0.943 \\
    \textbf{V4} &  &  &  &  & 1.0 \\
    \bottomrule
    \end{tabular}
    }
    \label{tab:FN_correlation}
\end{wraptable}
Ordering the scenes with respect to the number of FN, it becomes evident, that across all the datasets, specific scenes almost always result in FN results. In contrast, the majority of scenes have no FN results at all. Representing the number of FN per scene in a vector, we can compute the Pearson correlation between all dataset combinations. \Cref{tab:FN_correlation} clearly illustrates the high correlation between the datasets, i.e.\ the identical scenes across different datasets tend to produce FN results. Additionally, a less pronounced but still notable correlation to the real dataset is given.

\begin{figure}[tb]
    \centering

    \includegraphics[width=0.49\textwidth]{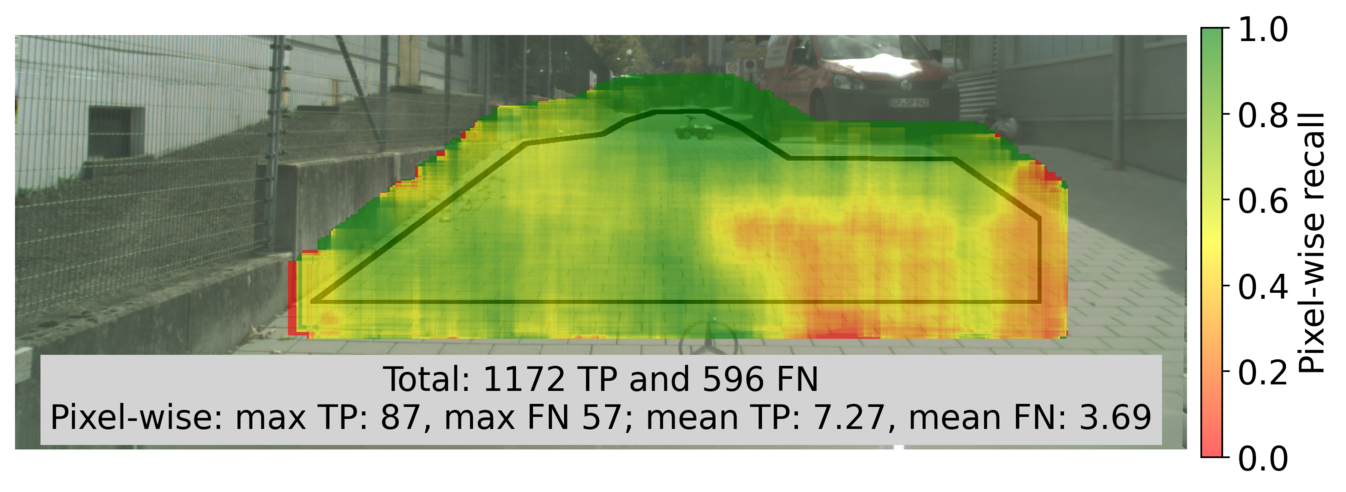}%
    \hfill
    \includegraphics[width=0.49\textwidth]{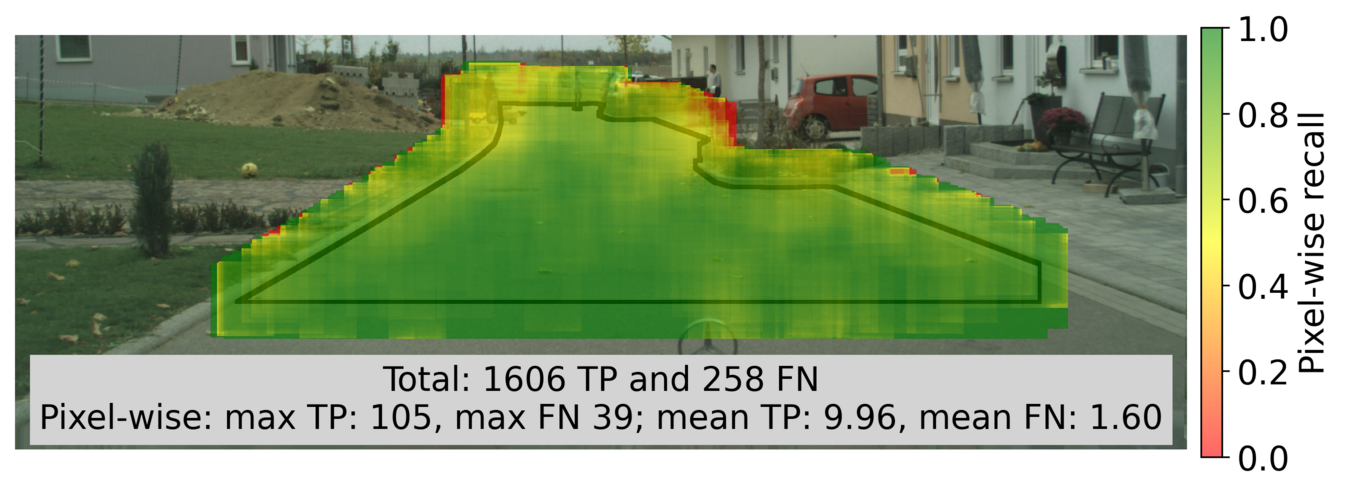}%

    \par\medskip

    \includegraphics[width=0.49\textwidth]{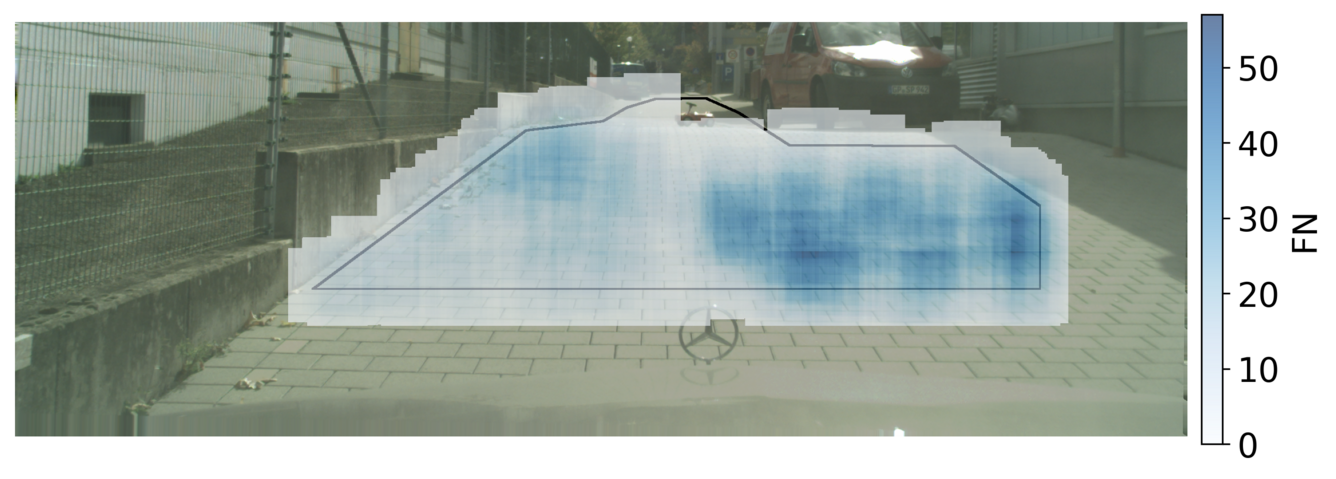}%
    \hfill
    \includegraphics[width=0.49\textwidth]{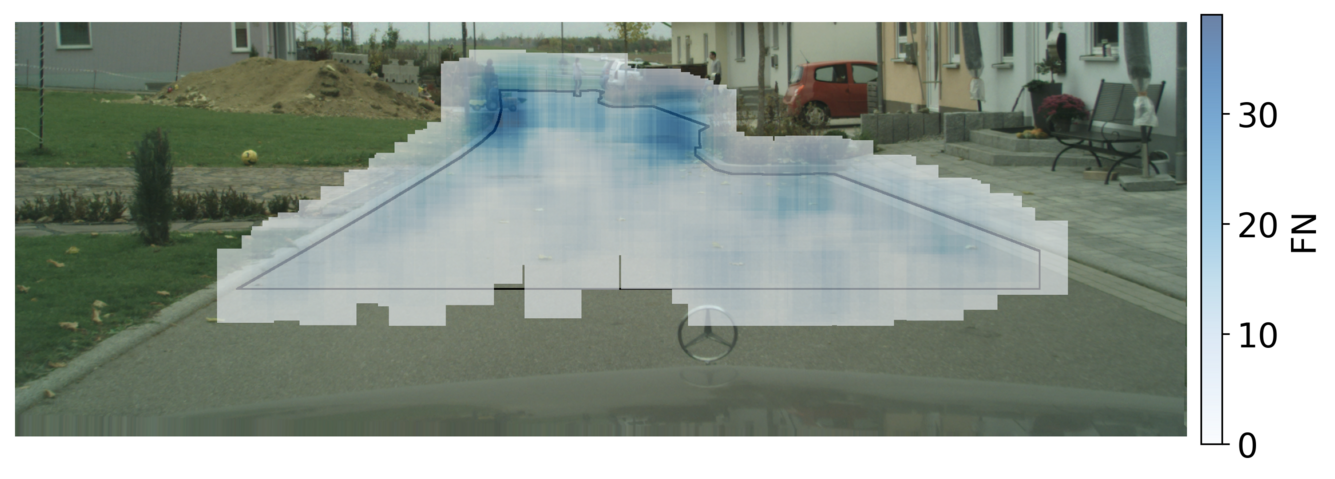}%

    \par\medskip

    \includegraphics[width=0.49\textwidth]{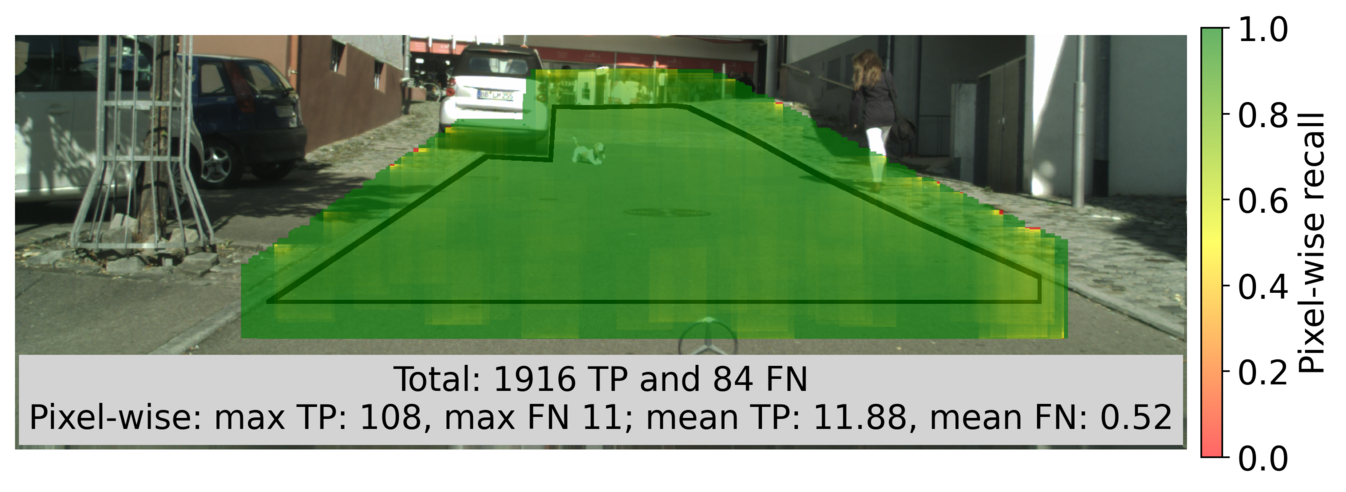}%
    \hfill
    \includegraphics[width=0.49\textwidth]{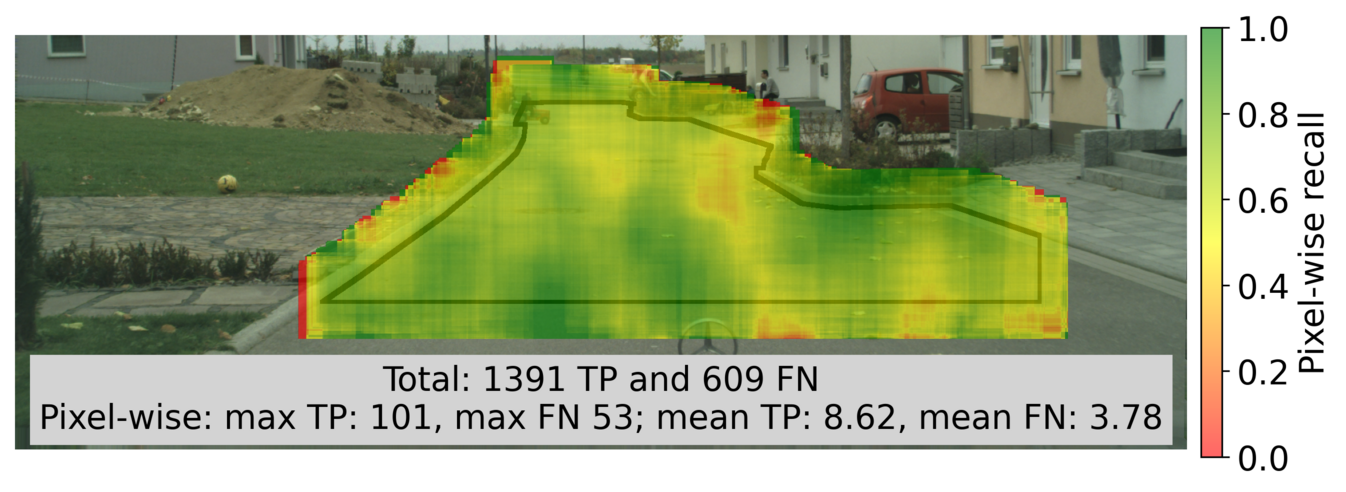}%

    \par\medskip

    \begin{minipage}[b]{0.49\textwidth}
        \centering
        \includegraphics[width=\textwidth]{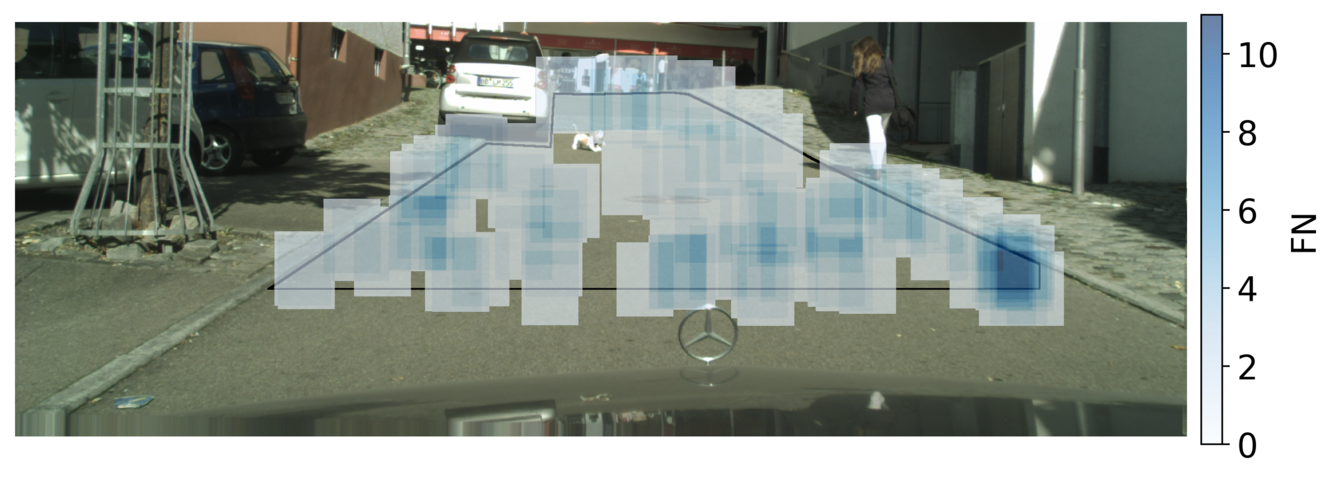}
        \caption*{(a) Few FN in fixed loc.}
        \label{fig:heatmaps_good_scene}
    \end{minipage}%
    \hfill
    \begin{minipage}[b]{0.49\textwidth}
        \centering
        \includegraphics[width=\textwidth]{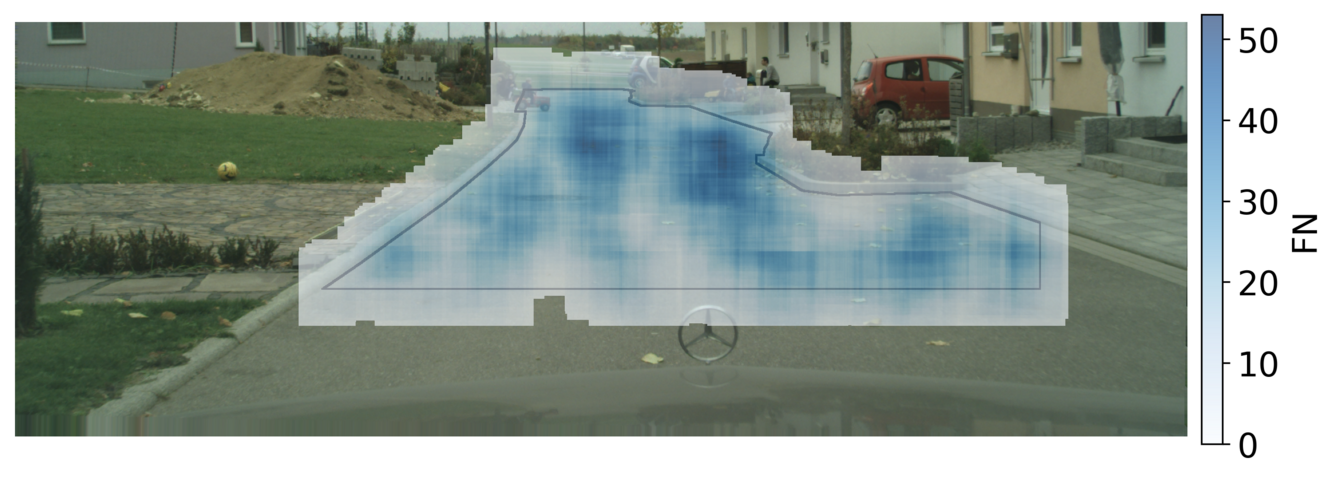}
        \caption*{(b) Many FN in fixed loc.}
        \label{fig:heatmaps_bad_scene}
    \end{minipage}

    \caption{Heatmaps of pixel-wise recall (first and second row) for Grounding DINO tested on inpainted objects at random locations in the street for scenes with (a) a low number of FN and (b) a high number of FN in our fixed location experiments in \cref{tab:LF4}. Heatmaps in the second and fourth row depict the number of FN corresponding to the above recall heatmap.}
    \label{fig:heatmap_good_bad_scene}
\end{figure}

\paragraph{Model performance across image locations.}
As the same scene across all datasets shows different inpainted objects, but all at the same location, we suspected the location within the scene plays a crucial factor in the performance.
We selected scenes that exhibited a high number of FN as well as those where most inpainted objects were detected. Our goal is to analyze how these detection outcomes change when objects need to be detected in different locations within the same scenes.
Different than before, we now inpaint objects based on random prompts to random locations on the road rather than replacing the real object of images from LostAndFound. To this end, all ground truth pixels annotated with road or object serve as potential bbox centers. 
We have created a bbox of $100 \times 130$ pixel for each chosen center pixel accordingly and used this to inpaint into a copy of the original image. In this way, we obtained 2,000 images with random objects for each selected street scene. Additional technical details are provided in appendix C.

To visualize the quality of the prediction results, we calculate the pixel-wise recall with a score threshold of $0.2$ and additionally provide a heatmap showing per pixel how often objects were overlooked. Note that, we see extreme recall values (dark green, i.e.\ recall of $1$, or dark red, i.e.\ recall of $0$) more often at the outer border due to the low sampling rate at those pixels. In \cref{fig:heatmaps_good_scene} we show two examples in which objects (inpainted at the location of the original LostAndFound object) were often successfully detected by Grounding DINO. However, we observe false negative clusters in different areas of the image. For the upper image, objects in the lower right corner are extremely difficult to detect, as well as objects located in the upper left area are more often overlooked. In addition to pixel clusters of FN detections, the second example in \cref{fig:heatmaps_good_scene} is showing multiple areas with very high recall, i.e.\ in specific areas where objects were never overlooked as depicted in the FN heatmaps. Similarly, scenes in which most inpainted objects were overlooked at the original location, see \cref{fig:heatmaps_bad_scene}, show high pixel-wise recall areas further away from the original LostAndFound object. These observations suggest that there is no specific scene that is the most challenging for Grounding DINO but rather the object location within the scene determines the difficulty.
This finding generalizes across models and to the different prompt types (see appendix C for exemplary visualizations). Furthermore, the areas where Grounding DINO fails vary across prompts and models. This further emphasizes the importance of prompt selection when applying open-vocabulary object detectors in safety critical applications.

\begin{figure}
    \centering
    \includegraphics[width=0.49\linewidth]{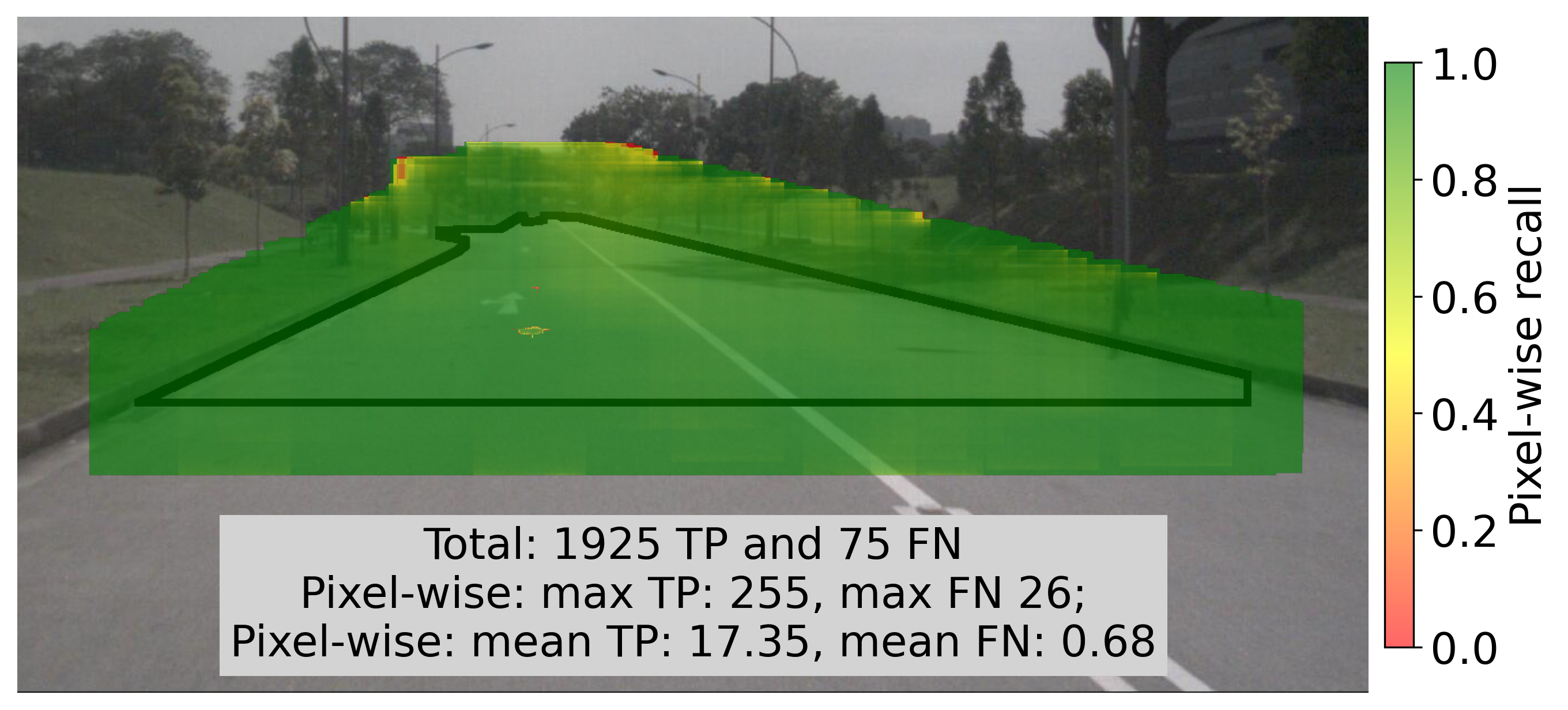}
    \includegraphics[width=0.49\linewidth]{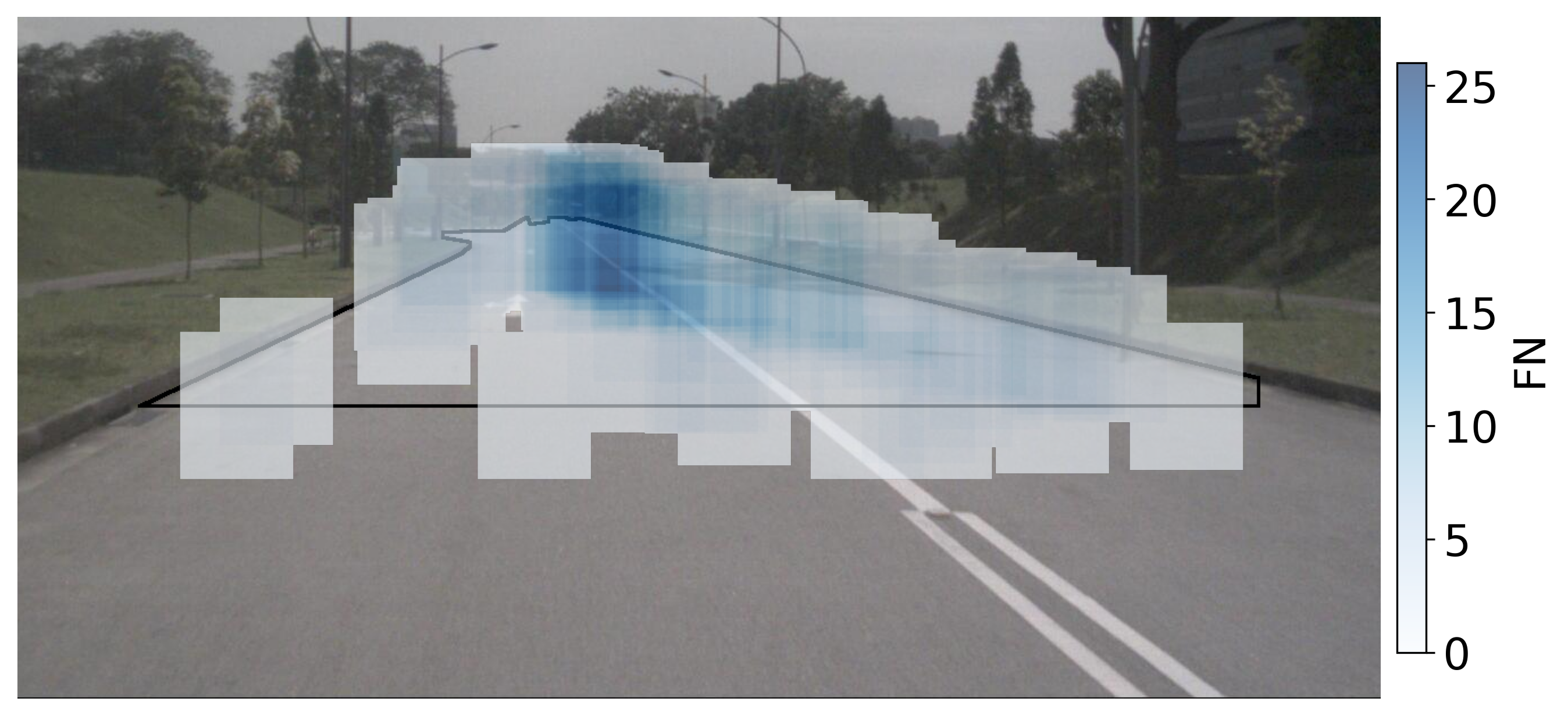}
    \caption{Heatmaps of the prediction performance of Grounding DINO given the prompt ``object on the street'' when detecting a ``robot'', synthetically generated and inpainted, at random locations on the road of an image of NuImages.}
    \label{fig:heatmaps_nuimage}
\end{figure}

To investigate in more depth the hypothesis that overlooking objects depends more on the location rather than their semantics and to show that this finding generalizes to the NuImages dataset, we use the fixed object prompt ``robot'' to inpaint at random locations on the street as was done for LostAndFound. 
The experimental results show that even though the object semantic is consistent, we still observe clusters of FN predictions, as exemplary demonstrated in \cref{fig:heatmaps_nuimage}. This indicates a blind spot in the model's detection capability,  i.e.\ a region in the image where it consistently overlooks objects. However, the location of this blind spot varies across scenes as already observed for LostAndFound.

\section{Conclusion}
    In this work, we investigated whether open-vocabulary object detectors can be systematically challenged with synthetically generated and inpainted objects. With the help of two inpainting methods, both based on stable diffusion, we inpainted either abnormal objects by randomly combining substantives from WordNet or sampling from a fixed prompt list of unusual but realistically looking objects for street scenes. This allows for a broad systematic exploration. 
Our experiments showed that all object detectors under test overlook objects -- real as well as synthetic ones -- on typical street scenes. In particular, the prompt used for open-vocabulary models as well as specific scenes seem to have a more significant influence on detections rather than the semantics of the objects. Moreover, the object location seems to have a high influence on the detection capability of the detectors. Our in-depth study on NuImages reveals that even though the object semantic is consistent, we observe clusters of FN predictions, i.e.\ regions in the image where it consistently overlooks objects, when randomly positioning objects on the road. In future work, we plan to investigate whether these findings can now be exploited to create challenging real-world datasets by putting real-world objects at the systematic failure areas and improving the models by fine-tuning on either dedicated synthetic failure cases or on the challenging real-world data.

\section*{Acknowledgment}
A.M.\ and M.R.\ acknowledge support by the German Federal Ministry
of Education and Research (BMBF) within the junior research group project ``UnrEAL'' (grant no.\ 01IS22069). Furthermore, M.R.\ acknowledges funding by the BMBF project ``REFRAME'' (grant no.\ 01IS24073C).

\includegraphics[width=4cm]{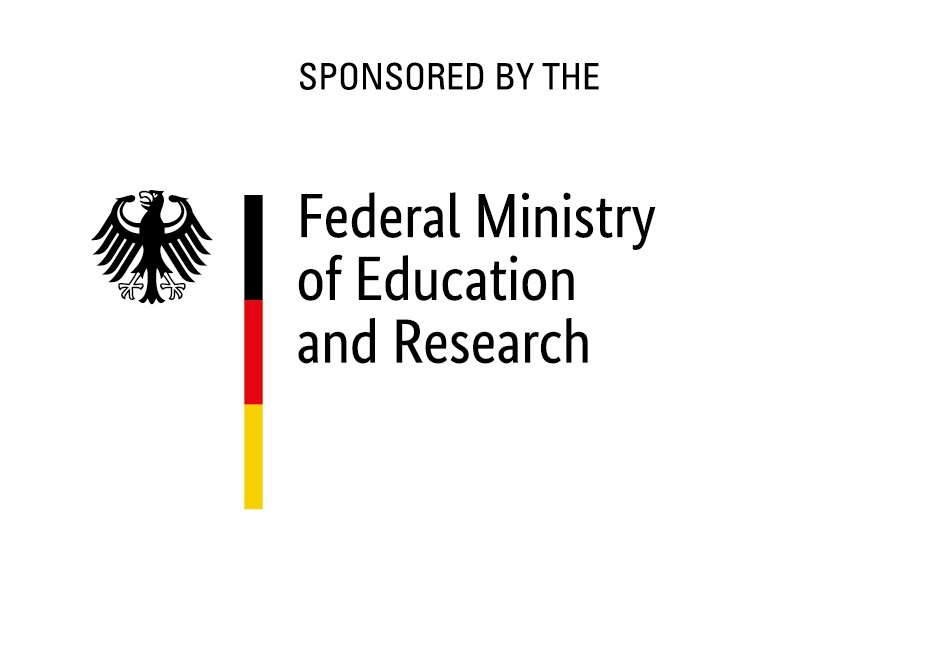}

\bibliographystyle{myrecomb}

\bibliography{bib}

\newpage
\appendix
    In the appendix, we provide technical details as well as additional visualizations for the discussions in the main paper.
\section{Technical and implementation details}

\subsection{Models}
We start with a short overview of all models used in our experiments. In \cref{tab:models}, we list for all models their specification and the datasets they were trained on.

\textbf{Grounding DINO} \cite{liu2023grounding} is formed on a Transformer-based detector DINO \cite{zhang2022dino} enabling easy processing of both text and image data. Combining this architecture with grounded pre-training makes Grounding Dino an innovative open-vocabulary model, allowing the detection of various objects based on human prompts. The model consists of three main components: feature enhancer, language-guide query selection, and cross-modality decoder. 
First, multi-scale features are extracted using Swin Transformer \cite{liu2021swin} as a backbone for image features and BERT \cite{devlin2018bert} for text features. Vanilla features are then fed into the feature enhancer where self-attention along with image-to-text and text-to-image cross-attention are applied. 
For effectively using the prompt to steer object detection, the language-guided query selection module chooses out of the processed features the ones that are more relevant to the input text as decoder queries. 
In the cross-modality decoder, both image and text modalities are merged by employing self-attention, cross-attention using the text and image features and a feed-forward neural network.

\textbf{YOLO-World} \cite{cheng2024yolo} is an open-set model that extends the well-known YOLO object detection framework \cite{redmon2016you} by incorporating open-vocabulary detection through vision-language modeling.
The open-set capabilities are based on a new re-parameterizable Vision-Language Path Aggregation Network (RepVL-PAN) and region-text contrastive loss to learn interaction between region-text pairs and thus visual and linguistic information. 
Other parts of the architecture are a YOLOv8 \cite{Jocher_Ultralytics_YOLO_2023} backbone for image feature extraction and CLIP \cite{radford2021learning} as a model to convert the nouns of a given prompt into embeddings. Then both feature representations are fed to the RepVL-PAN and fused at multiple levels. Ultimately a text-contrastive head regresses bounding boxes and object embeddings.

\textbf{MDETR} \cite{kamath2021mdetr} is an end-to-end text-modulated detection method derived from DETR \cite{carion2020end}. It utilizes a convolutional backbone for visual feature extraction and a language model, RoBERTa \cite{liu2019roberta}, for text feature extraction. 

\textbf{OmDet-Turbo} \cite{zhao2024real} introduces an Efficient Fusion Head (EFH) module, aiming for real-time performance. The overall OmDet-Tubro model features a text backbone, an image backbone and an EFH module. EFH includes an Efficient Language-Aware Encoder (ELA-Encoder) and Decoder (ELA-Decoder). ELA-Encoder selects top-K initial queries, fused with prompt embedding for language-guided multi-modality queries in ELA-Decoder. 

We use the above-mentioned open-vocabulary models by providing them with different text prompts and explore to what extent they could be utilized for object detection on synthetic data. We also use a classical closed-vocabulary object detection model, FasterRCNN \cite{ren2016fasterrcnnrealtimeobject}, as reference to compare against the open-vocabulary models.

\begin{table}[htb]
\centering
\scriptsize
  \caption{Information about the models belonging grouped according to different categories i.e., open-vocabulary and classical (object) detection. It also specifies the backbone and pre-trained data used for the models in our experiments.}
  \resizebox{0.8\textwidth}{!}{

  \begin{tabular}{@{}llll@{}}
    \toprule
    \textbf{Category} & \textbf{Model} & \textbf{Backbone} & \textbf{Pre-trained Data}\\
    \midrule
    open-vocabulary & YOLO-World-L & YOLOv8-L & O365,GoldG\\
    & Grounding DINO-T & Swin-T & O365,GoldG,Cap4M\\
    & MDETR & EfficientnetB5 & VG,COCO,Flickr30k\\
    & OmDet-Turbo & Swin-T & O365,GoldG\\
    \midrule
    class. detection & FasterRCNN & ResNet50 & COCO  \\
    \bottomrule
  \end{tabular}
  }
    \label{tab:models}
\end{table}

\subsection{Datasets and inpainting}
Inpainting includes setting various parameters that influence the generation or inpainting process. These are for example the sampling method which is used to accelerate the denoising process. Over time various approaches emerged \cite{lu2022dpm, song2020denoising}. Therefore, choosing the right sampler can highly influence the inpainting results. We used the API mode to process a large amount of images.
To analyze the scene dependency, we varied the hyperparameters of the inpainting method slightly and created three additional dataset (LostAndFound V1 - LostAndFound V3). The quantitative results shown in the main paper are based on LostAndFound V4.
The hyperparameters used to create the datasets are listed in \cref{tab:dataset}.  We adapted the sampler, denoising strength, and inpainting fill parameters. For the other important settings, we always stuck with the default parameters of the stable diffusion-based dreamshaper\_8inpainting model \cite{-_2023} and a number of 30 sampling steps. 
\begin{table}[htb]
\centering
\scriptsize
  \caption{Information about the inpainted datasets created for our experiments. It also specifies the parameters used for the inpainting process.}
  \label{tab:dataset}
  \resizebox{0.8\textwidth}{!}{
  \begin{tabular}{@{}llll@{}}
    \toprule
    \textbf{Dataset Name} & \textbf{Sampler Name} & \textbf{Denoising Strength} & \textbf{Inpainting Fill}\\
    \midrule
    LostAndFound V2  & Euler a & 0.7 & False  \\
    LostAndFound V3  & DPM++ 2S a & 0.7 & False  \\
    LostAndFound V4  & DPM++ 2S a & 1 & False  \\
    \bottomrule
  \end{tabular}
  }
\end{table}

\section{Model performance across datasets}
In addition to the results shown in the main paper, we computed the performance of the 5 models when tested on LostAndFound and NuImages using the single-concept inpainting. Results are listed in \cref{tab:LF_SI}, where we compare the results in four different settings. The Single-concept inpaintings resulted usually into smaller sized objects, that made it difficult to evaluate due to the discrepency with the ground truth bounding boxes. Therefore, to make the results comparable we reduced the IoU threshold to 0.1. The results conclude that both inpainting procedures were equally challenging for the models, however, NuImages w/ single-concept inpainting has the highest AUPRC. That suggests that the quality of the inpaintings did alter the performance of the models. In general GroundingDINO performs the best with an AUPRC of 90\%, but MDETR is fairly better in terms of the AP score of 24.9\% particularly on NuImages. 

\begin{table}
\caption{Comparison of results for 5 different models on synthetic LostAndFound and NuImages using single-concept inpainting.}
\label{tab:LF_SI}
\setlength{\tabcolsep}{4pt}

\resizebox{\textwidth}{!}{
\begin{tabular}{ll*{12}{c}}
\toprule
\multicolumn{2}{c}{} 
& \multicolumn{3}{c}{\parbox[c]{4cm}{\centering \textbf{LostAndFound w/ single-concept inpainting}}}
& \multicolumn{3}{c}{\parbox[c]{4cm}{\centering \textbf{NuImages w/ single-concept inpainting}}}
& \multicolumn{3}{c}{\parbox[c]{4cm}{\centering \textbf{LostAndFound w/ Hybrid-concept inpainting}}}
& \multicolumn{3}{c}{\centering \textbf{LostAndFound }} \\
\cmidrule(lr){3-5} \cmidrule(lr){6-8} \cmidrule(lr){9-11} \cmidrule(lr){12-14}
\textbf{Model} & \textbf{Prompt} 
& \textbf{AUPRC} & \textbf{AP} & \textbf{AR} 
& \textbf{AUPRC} & \textbf{AP} & \textbf{AR}
& \textbf{AUPRC} & \textbf{AP} & \textbf{AR}
& \textbf{AUPRC} & \textbf{AP} & \textbf{AR} \\
\midrule
\multirow{5}{*}{\rotatebox[origin=c]{90}{MDETR}} 
 & p1 & 0.2480 & 0.0482 & 0.2637 
  & 0.6519 & \textbf{0.2495} & 0.3261 
  & 0.2888 & 0.0506 & 0.1980 
 & 0.2205 & 0.1262 & 0.4335 \\
 & p2 & 0.3842 & 0.0841 & 0.2889  
 & 0.4349 & 0.2489 & 0.3591  
 & 0.3819 & 0.0455 & 0.1862 
& 0.4117 & 0.1714 & 0.4291 \\
 & p3 & 0.2399 & 0.0589 & 0.3080  
  & 0.6684 & 0.2453 & 0.3311  
  & 0.2835 & 0.0402 & 0.2073
& 0.2450 & 0.1173 & 0.4464 \\
 & p4 & 0.1126 & 0.0322 & 0.2459  
  & 0.5238 & 0.2146 & 0.2935  
  & 0.1486 & 0.0282 & 0.1859 
 & 0.1821 & 0.1139 & 0.3453 \\
 & p5 & 0.1668 & 0.0466 & 0.2566  
  & 0.4938 & 0.1847 & 0.2477  
  & 0.1776 & 0.0270 & 0.1693
 & 0.1415 & 0.0665 & 0.3229 \\

\midrule
\multirow{5}{*}{\rotatebox[origin=c]{90}{OmDet}}

& p1 & 0.4323 & 0.0892 & 0.3012
& 0.6862 & 0.1085 & 0.2480 
& 0.4821 & 0.1221 & 0.3012
& 0.3859 & 0.2945 & 0.7101  \\

& p2 & 0.2274 & 0.0507 & 0.3488
& 0.5055 & 0.1885 & 0.2806
& 0.1326 & 0.0390 & 0.3226
& 0.2114 & 0.1617 & 0.7860  \\
& p3 & 0.4399 & 0.1220 & 0.3498
& 0.7030 & 0.1284 & 0.2700
& 0.4802 & 0.0949 & 0.2503
&0.2861 & 0.1947 & 0.4978  \\
& p4 & 0.2181 & 0.0566 & 0.3326
& 0.5171 & 0.1051 & 0.3143
& 0.2136 & 0.0246 & 0.1827
&0.1709 & 0.0733 & 0.4341  \\
& p5 & 0.0202 & 0.0088 & 0.0919
& 0.0856 & 0.0064 & 0.0706 
& 0.0208 & 0.0058 & 0.0595
& 0.0237 & 0.0226 & 0.1592  \\

\midrule
\multirow{5}{*}{\rotatebox[origin=c]{90}{G. DINO}}
& p1 & \textbf{0.7795} & 0.1632 & 0.3732
& \textbf{0.9066} & 0.1477 & 0.3250
& \textbf{0.8063} & 0.1975 & 0.3319
& \textbf{0.6793} & 0.5488 & 0.7352 \\

& p2 & 0.1751 & 0.0438 & 0.3680
& 0.3823 & 0.0593 & 0.3109
& 0.1469 & 0.0477 & 0.3284
& 0.1928 & 0.1687 & 0.8196\\

& p3 & 0.3683 & 0.0847 & 0.3255
& 0.2362 & 0.0302 & 0.2496
& 0.5467 & 0.0793 & 0.2293
& 0.3843 & 0.2285 & 0.5654\\

& p4 & 0.0629 & 0.0191 & 0.2413
& 0.3098 & 0.0505 & 0.2805
& 0.1393 & 0.0192 & 0.1887
& 0.0814 & 0.0465 & 0.4034\\

& p5 & 0.1755 & 0.0142 & 0.1719 
& 0.7284 & 0.1076 & 0.3136
& 0.1983 & 0.0123 & 0.1251
& 0.1525 & 0.0264 & 0.2631\\

\midrule
\multirow{5}{*}{\rotatebox[origin=c]{90}{YOLO World}}
 & p1 & 0.3364 & 0.0740 & 0.1327 
 & 0.0132 & 0.0024 & 0.0033
& 0.1159 & 0.0355 & 0.0448
& 0.1173 & 0.1213 & 0.1173\\
 
 & p2 & 0.1155 & 0.0283 & 0.2757
 & 0.1963 & 0.0318 & 0.2457
& 0.0334 & 0.0132 & 0.1128
 & 0.0931 & 0.0753 & 0.4240\\

 & p3 & 0.6499 & 0.1175 & 0.2895
& 0.4176 & 0.0516 & 0.1335
 & 0.3723 & 0.0899 & 0.1984
 & 0.6754 & 0.4914 & 0.6531\\

 & p4 & 0.0226 & 0.0136 & 0.0124
& 0.0163 & 0.0094 & 0.0068
 & 0.0124 & 0.0082 & 0.0050
 & 0.0434 & 0.0548 & 0.0492\\

 & p5 & 0.0982 & 0.0327 & 0.1019 
& 0.1020 & 0.0259 & 0.0439
& 0.0245 & 0.0116 & 0.0381
& 0.0714 & 0.0749 & 0.2101\\

\midrule
\multirow{1}{*}{FRCNN}
 & - & 0.0870 & 0.0190 & 0.2680 
 & 0.1556 & 0.0223 & 0.2347 
 & 0.05842 & 0.0149 & 0.2529
 & 0.0329 & 0.0213 & 0.1659 \\
\bottomrule
\end{tabular}
}
\end{table}

\section{Model performance across image locations}
\begin{figure}[htb]
    \centering

    \begin{minipage}[b]{0.48\linewidth}
        \centering
        \includegraphics[width=\linewidth]{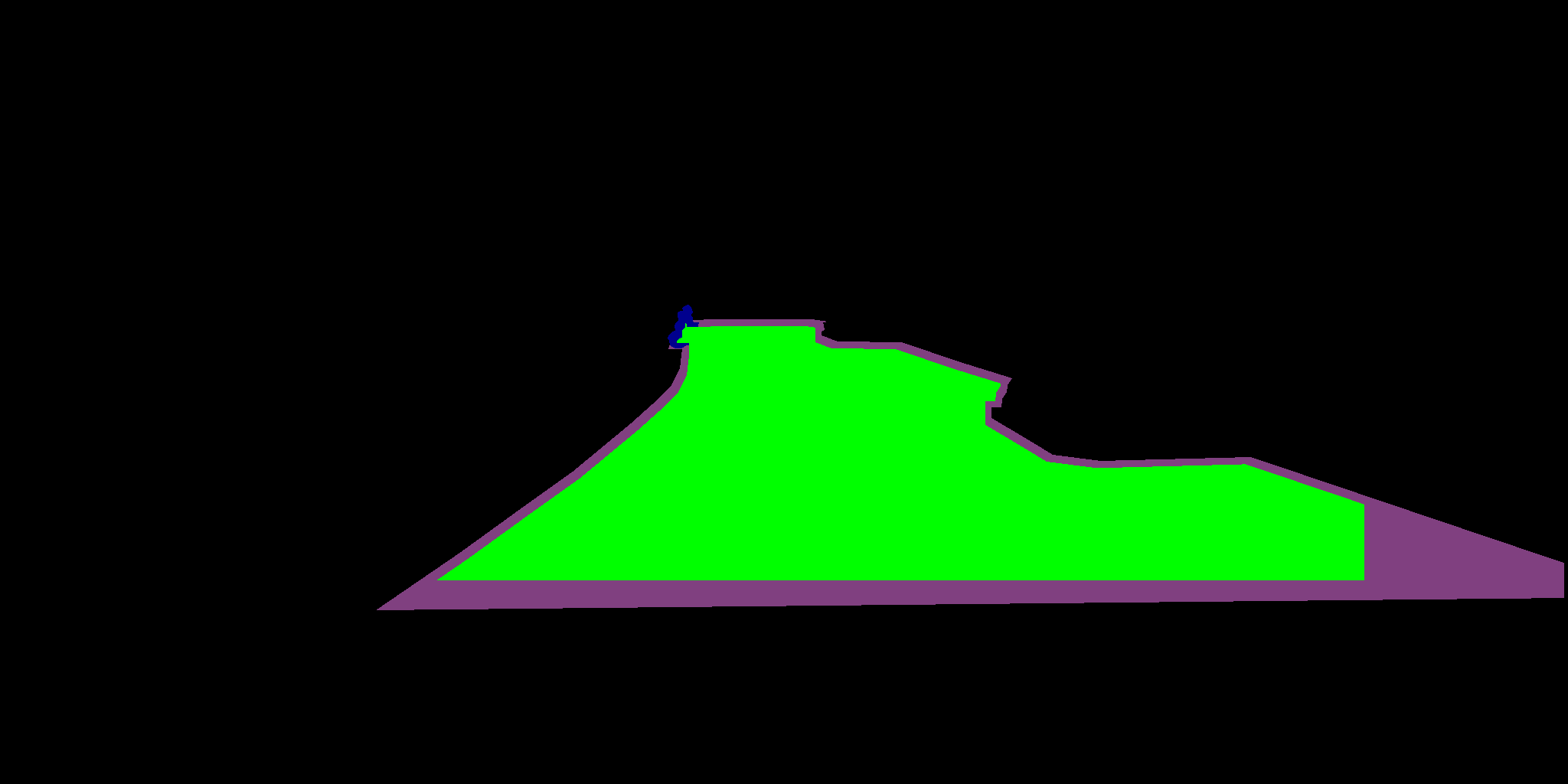}
        \caption*{(a) Road-only sample set (green)}
        \label{fig:a}
    \end{minipage}%
    \hfill
    \begin{minipage}[b]{0.48\linewidth}
        \centering
        \includegraphics[width=\linewidth]{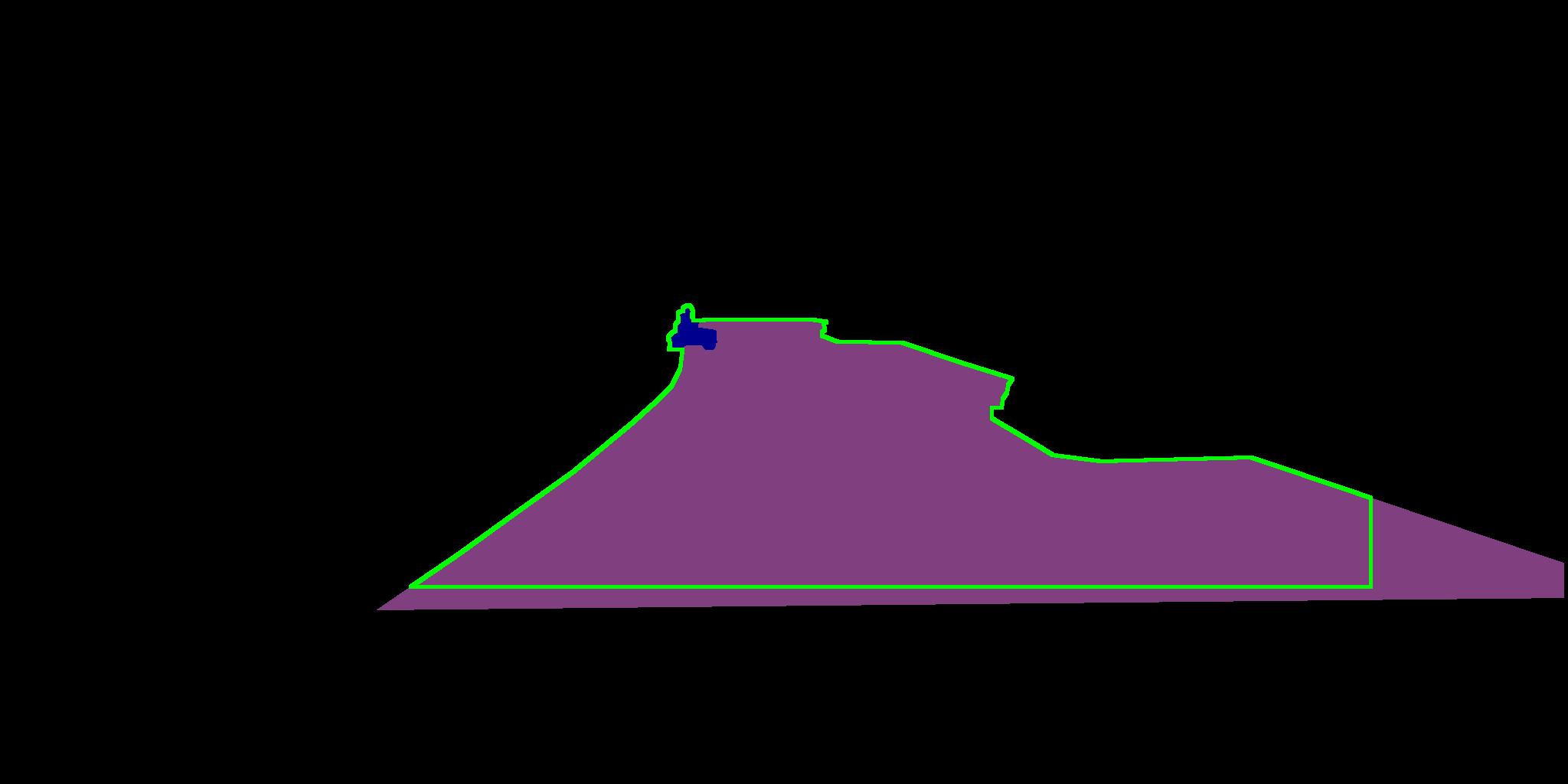}
        \caption*{(b) Border sample set (green)}
        \label{fig:b}
    \end{minipage}

    \caption{Based on the purple region of interest from the dataset ground truth, we created two sets of pixels (a) road pixels and (b) road border pixels, visualized in green. These are used to draw random centers for bboxes. To maintain our inpainting pipeline, we ensured that potential bbox center are always $512$ pixels away from the image border to allow smooth inpainting in the scene context.}
    \label{fig:border-vs-streetpixels}
\end{figure}

As described in the main paper, to investigate our hypothesis that the location of an object has a significant influence on the detection, we generate $2000$ copies of selected scene. In each copy of one scene, we inpaint one object with varying prompt at a random location of the road.
As shown in \cref{fig:border-vs-streetpixels}, pixels too close to the image border, making it impossible to retrieve a full $512\times 512$ frame, have been dismissed. Remaining pixels were split in a border and a road-only set. We then sampled random pixels, $1600$ from the road-only sample set and $400$ from the border set to obtain a broad coverage of the road area.

\begin{figure}[thb]
    \centering
        \begin{minipage}{0.49\textwidth}
            \centering
            \includegraphics[width=\textwidth]{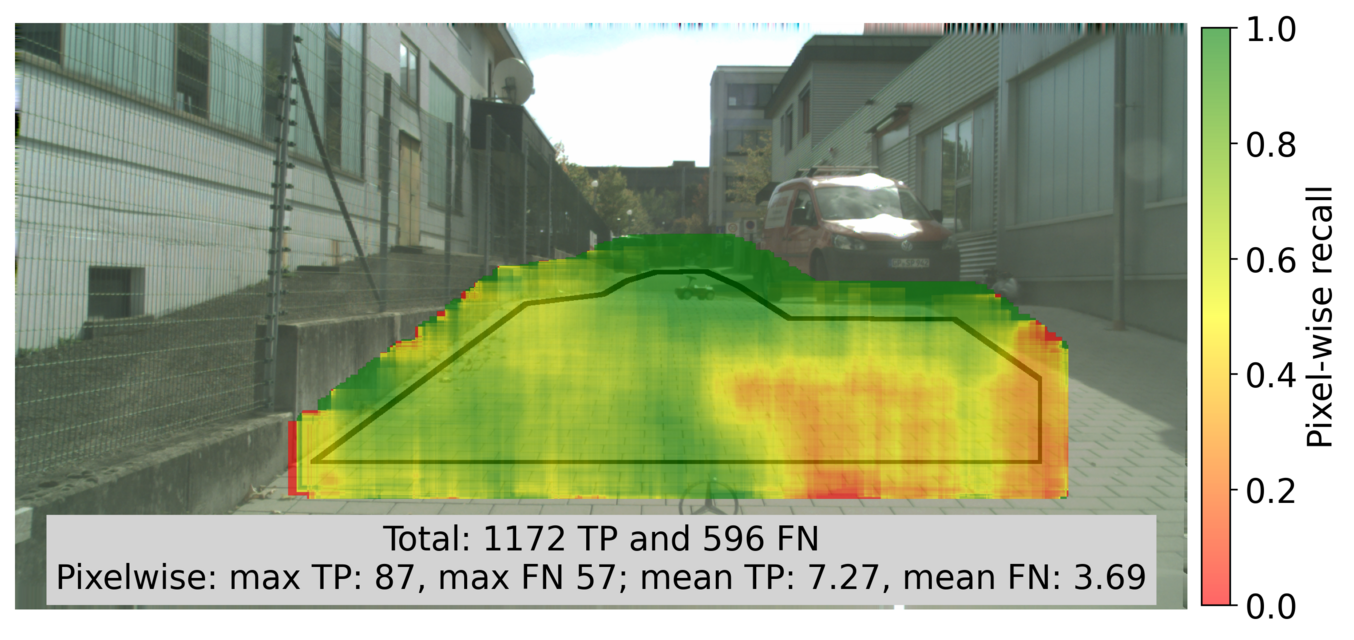}
        \end{minipage}
        \begin{minipage}{0.49\textwidth}
        \centering
        \includegraphics[width=\textwidth]{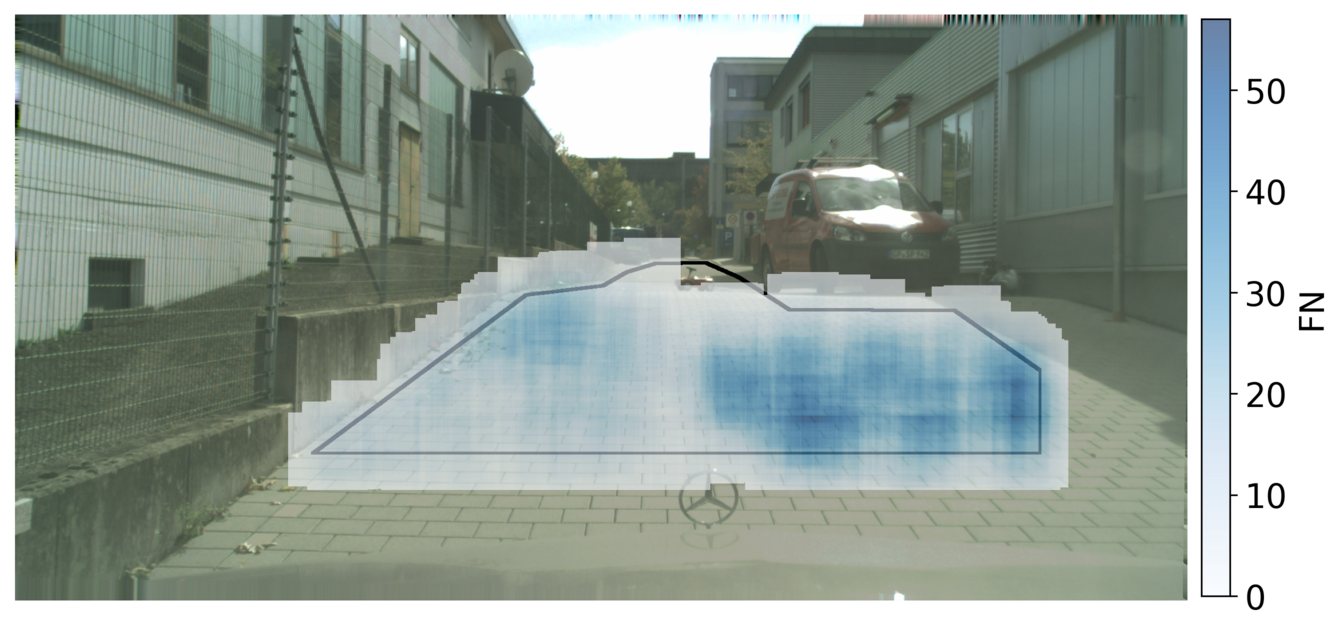}
        \end{minipage}
        \caption*{(a) Grounding DINO}
    \hfill
        \centering
        \begin{minipage}{0.49\textwidth}
            \centering
            \includegraphics[width=\textwidth]{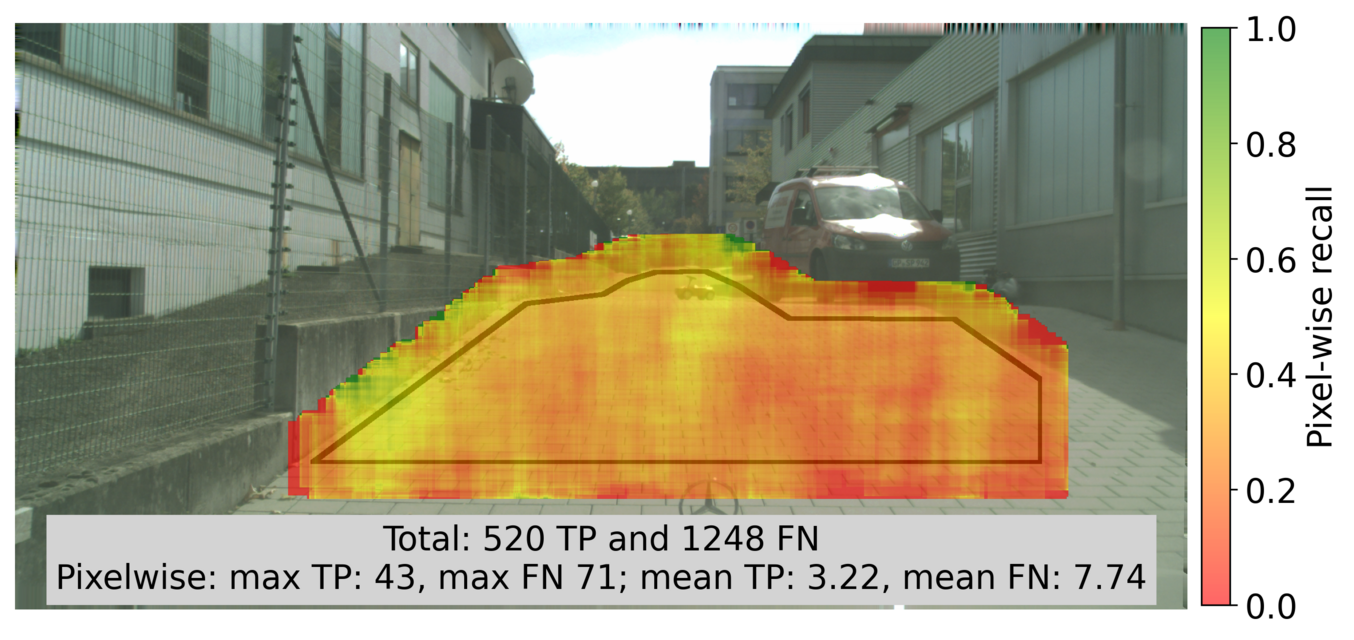}
        \end{minipage}
        \begin{minipage}{0.49\textwidth}
            \centering
            \includegraphics[width=\textwidth]{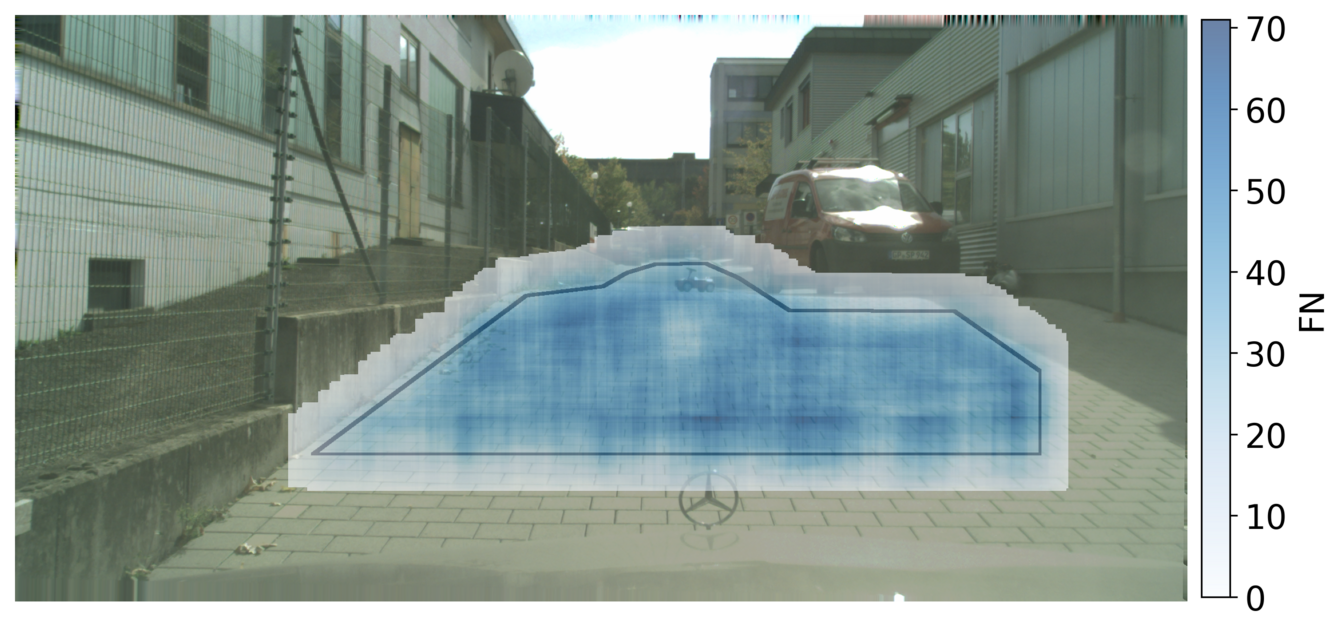}
        \end{minipage}
        \caption*{(b) YOLO-World} 
    \caption{Pixel-wise recall heatmaps for the same scene with varying models used for detection. The open vocabulary models Grounding DINO and YOLO-World were tested based on the prompt ``object on the street''.}
    \label{fig:heatmap_models_1}
\end{figure}

\begin{figure}[thb]
    \centering
        \begin{minipage}{0.49\textwidth}
            \centering
            \includegraphics[width=\textwidth]{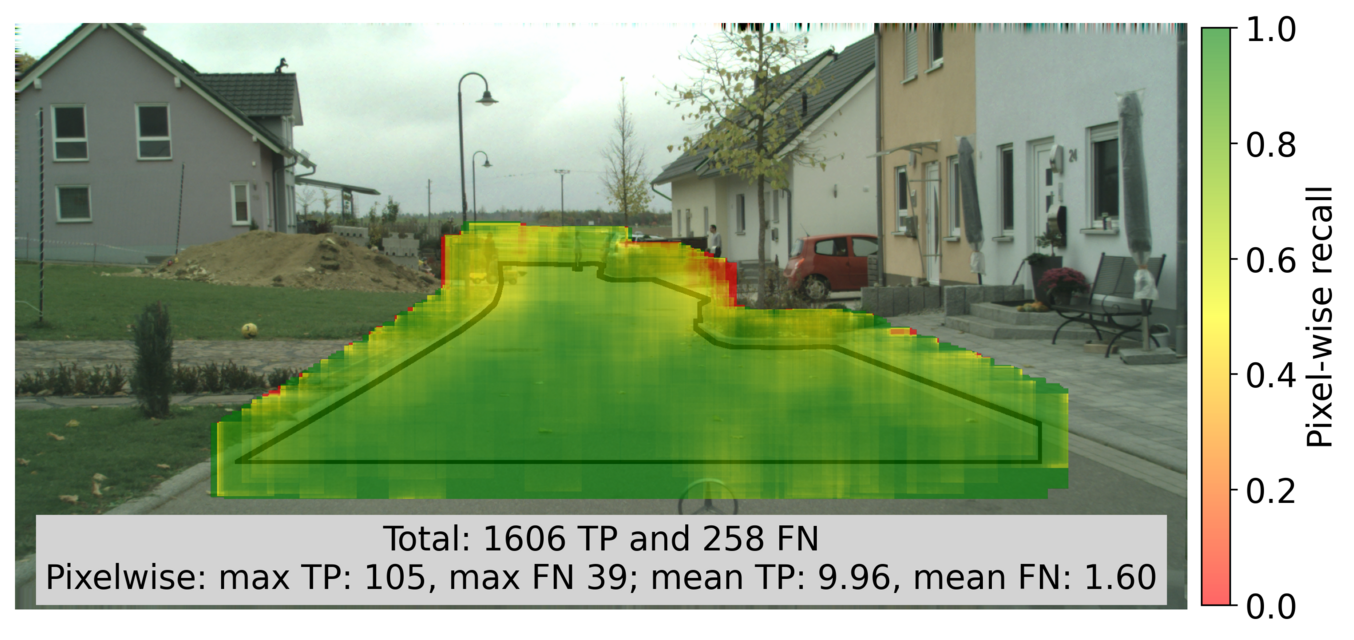}
        \end{minipage}
        \begin{minipage}{0.49\textwidth}
        \centering
        \includegraphics[width=\textwidth]{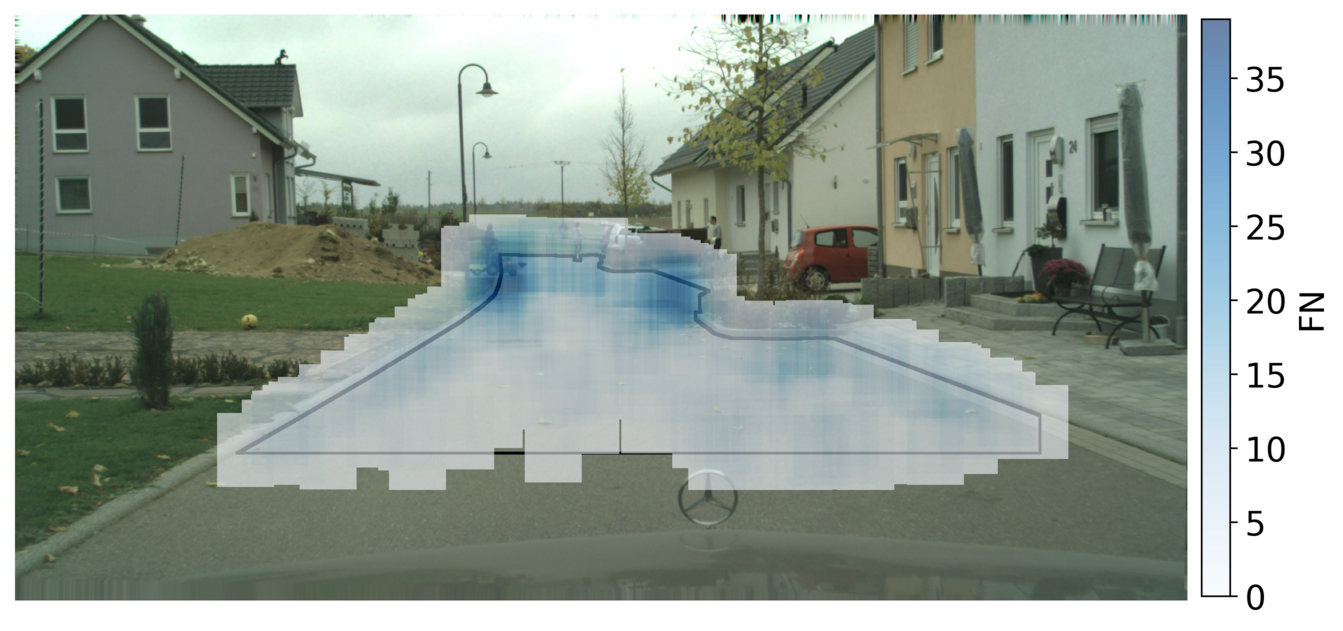}
        \end{minipage}
        \caption{Grounding DINO}
    \hfill
        \centering
        \begin{minipage}{0.49\textwidth}
            \centering
            \includegraphics[width=\textwidth]{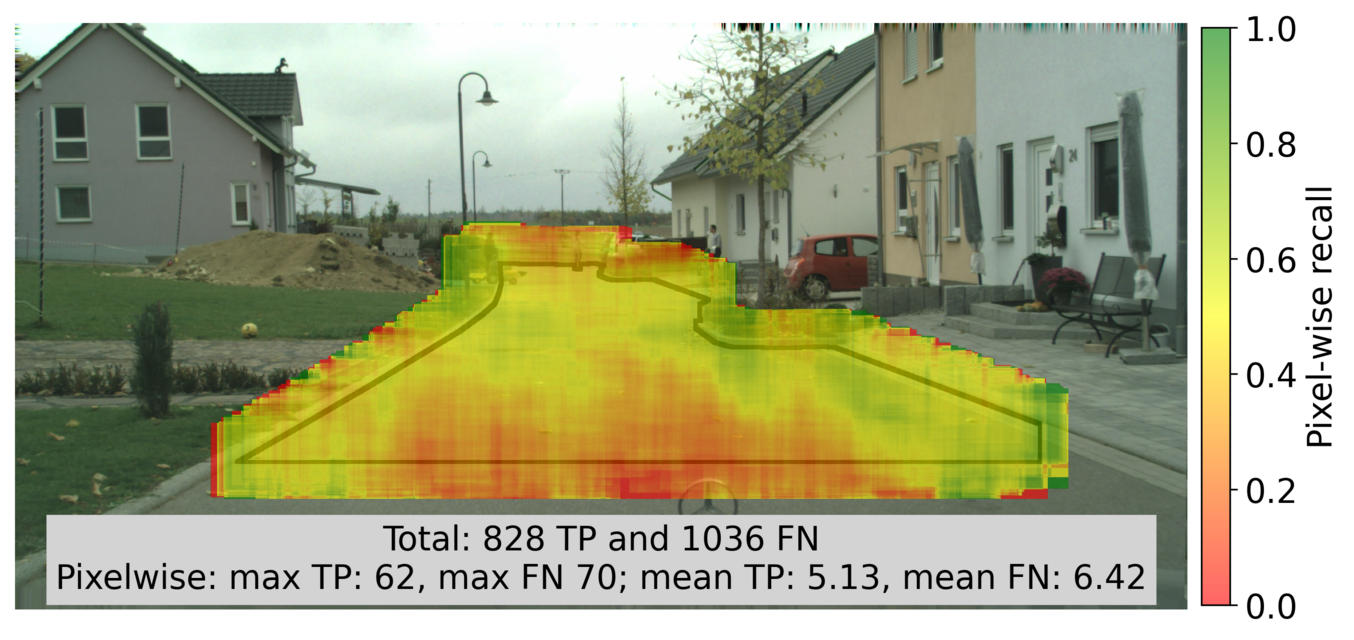}
        \end{minipage}
        \begin{minipage}{0.49\textwidth}
            \centering
            \includegraphics[width=\textwidth]{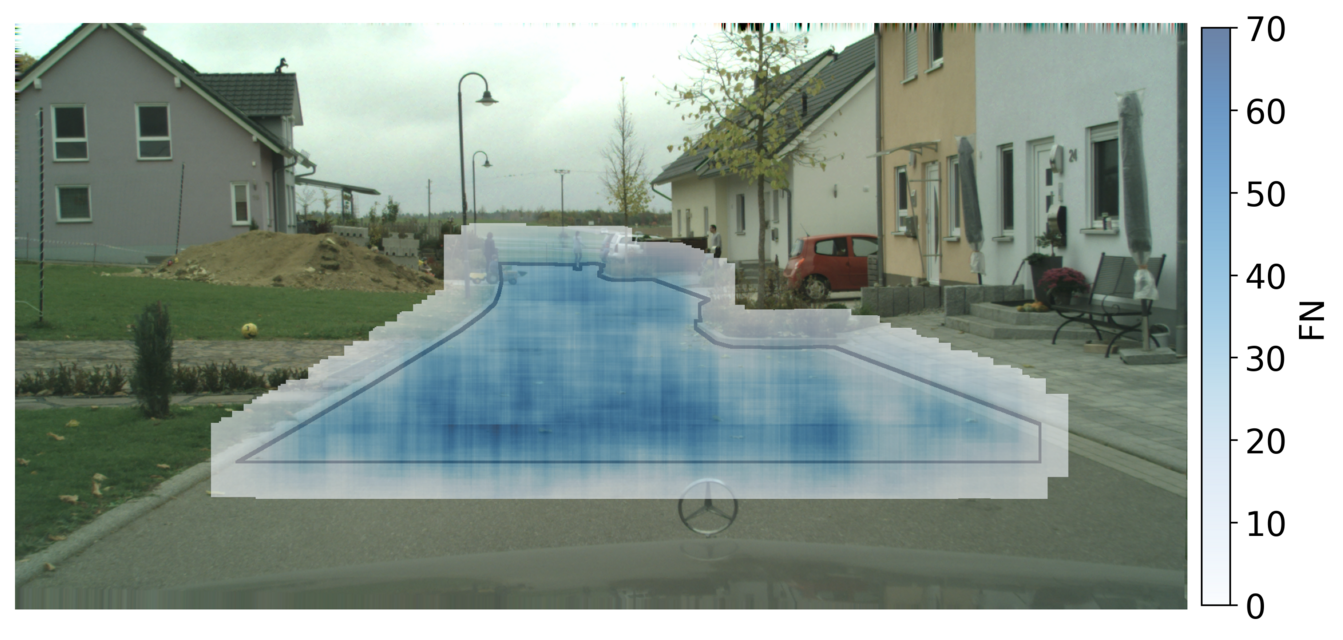}
        \end{minipage}
        \caption{YOLO-World}
    \caption{Pixel-wise recall heatmaps for the same scene with varying models used for detection. The open vocabulary models Grounding DINO and YOLO-World were tested based on the prompt ``object on the street''.}
    \label{fig:heatmap_models_2}
\end{figure}

When analyzing the prediction capability on random image locations across models, depicted exemplary in \cref{fig:heatmap_models_1,fig:heatmap_models_2}, we again observe FN clusters. However, those differ across the models. While YOLO-World can be challenged in most of the image regions, Grounding DINO has dedicated blind spots and performs more reliable on the remaining area. We fixed the prompt to test the models to ``object in the street'' in this experiment.

\begin{figure}[bt]
    \centering

    \begin{minipage}{0.49\textwidth}
        \centering
        \includegraphics[width=\textwidth]{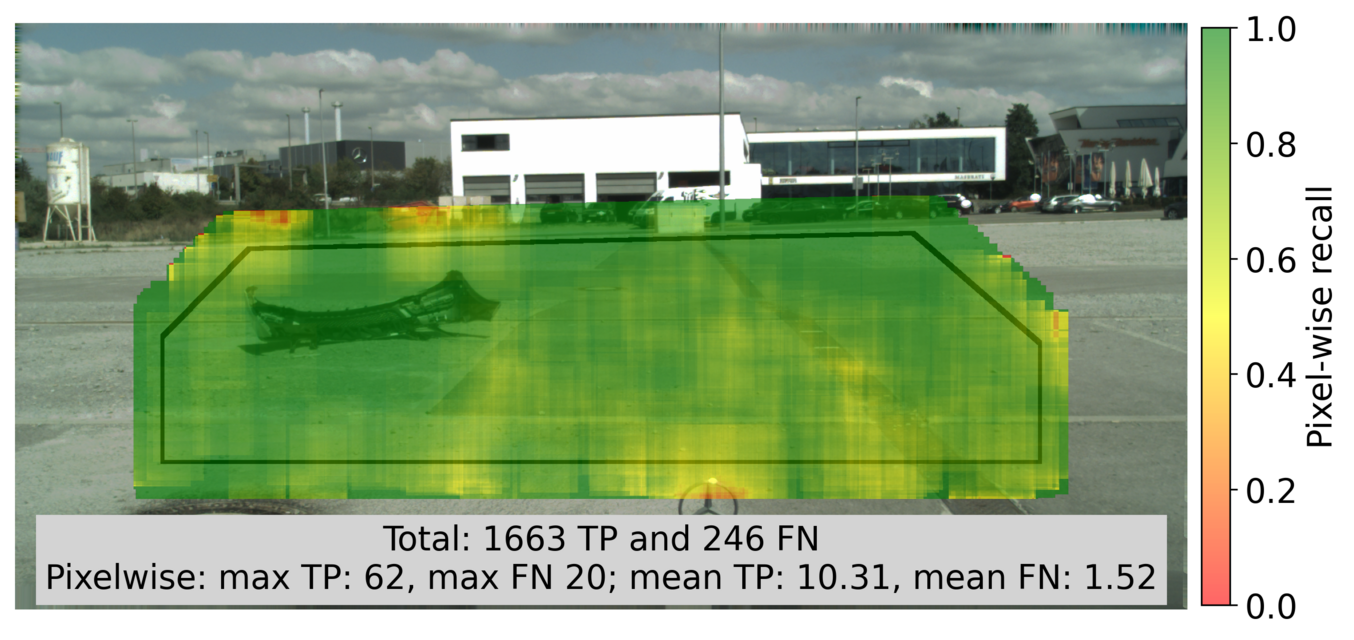}
    \end{minipage}%
    \hfill
    \begin{minipage}{0.49\textwidth}
        \centering
        \includegraphics[width=\textwidth]{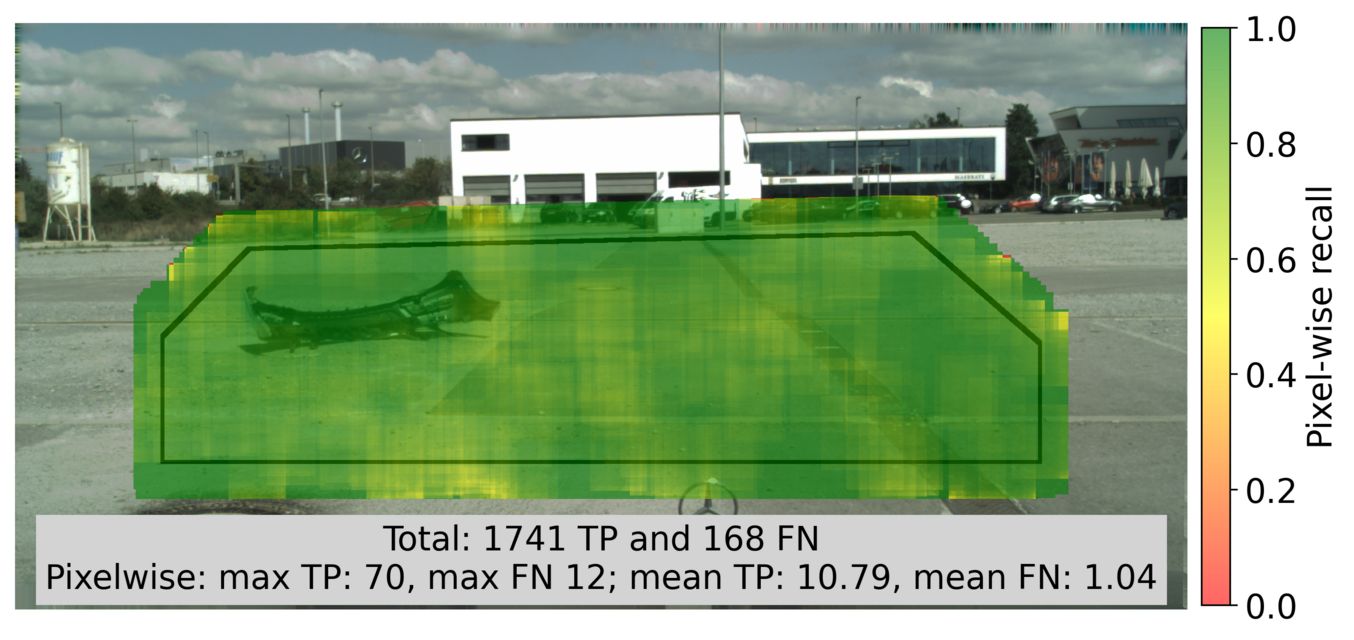}
    \end{minipage}

    \vspace{1em}

    \begin{minipage}{0.49\textwidth}
        \centering
        \includegraphics[width=\textwidth]{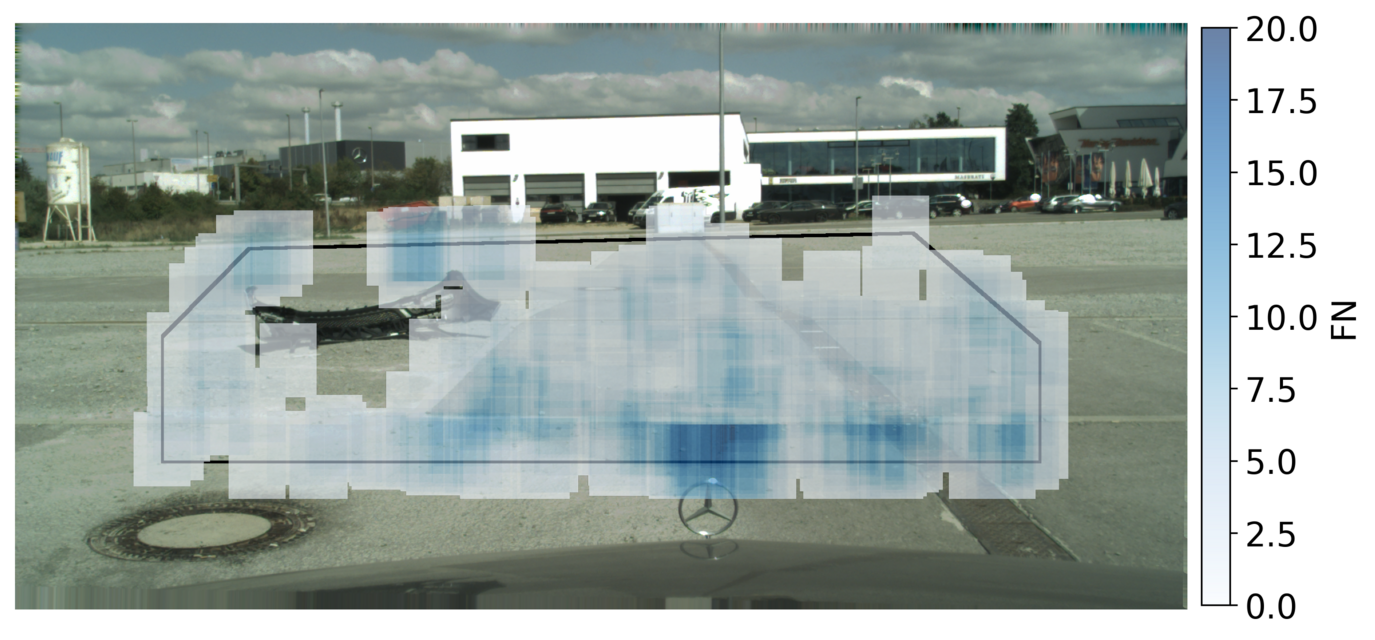}
        \caption*{Prompt p3: \textit{object on the street}}
    \end{minipage}%
    \hfill
    \begin{minipage}{0.49\textwidth}
        \centering
        \includegraphics[width=\textwidth]{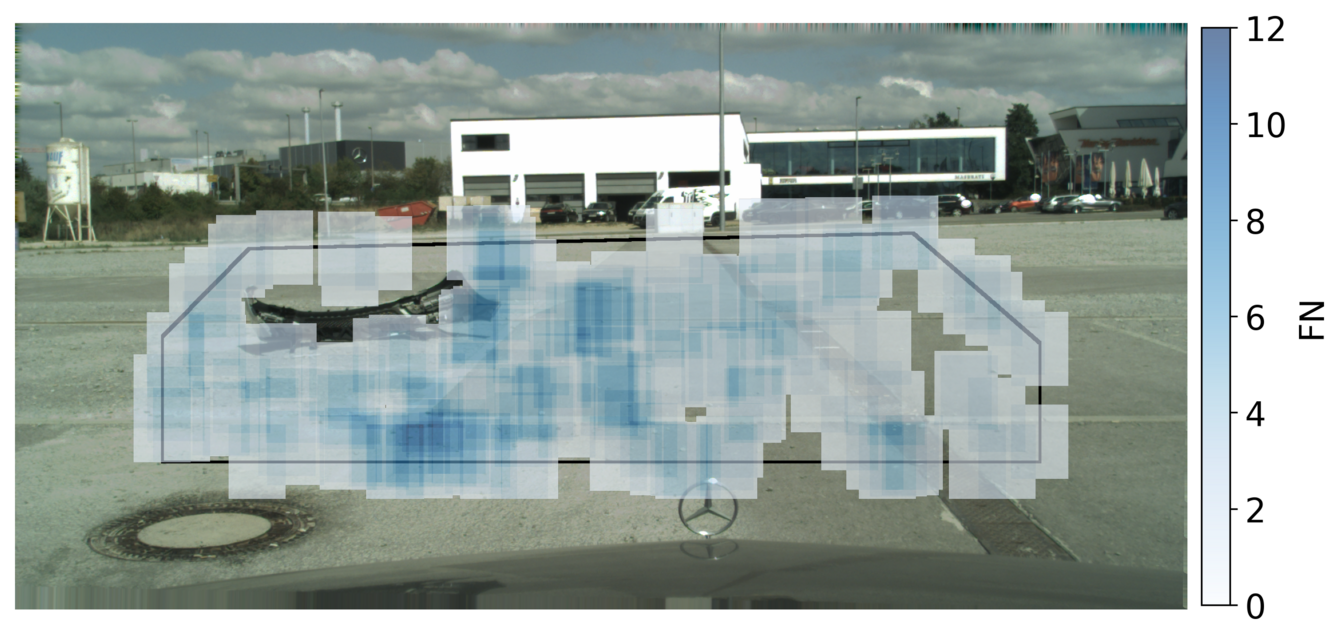}
        \caption*{Prompt p2: \textit{object . animal . person}}
    \end{minipage}

    \caption{Prediction quality of Grounding DINO with varying prompt. Heatmaps show pixel-wise recall as well as the number of FN for identic scenes from LostAndFound with different detection prompts.}
    \label{fig:heatmaps_prompt}
\end{figure}

\section{Additional experiments}
    Before and in between conducting the experiments in the main paper, we did some tests on a small image subset on similar detection tasks. This way, we wanted to rule out misleading errors or correlations. However, these experiments have not been realized on big sets of data. Therefore, we just explain them briefly in the following part.
    After our first detection tests on generated hybrids, we realized that synthetic content seemed to be relatively easy to detect for Grounding DINO. Therefore, we wanted to ensure that not the sake of synthetic content, rather than the actual appearance of objects or the underlying semantics results in TP detections. Considering that, we altered street scenes in different ways to see how this affects the detection performance.

\begin{figure}[h!]
    \centering

    \includegraphics[width=0.45\textwidth,height=5.5cm,keepaspectratio]{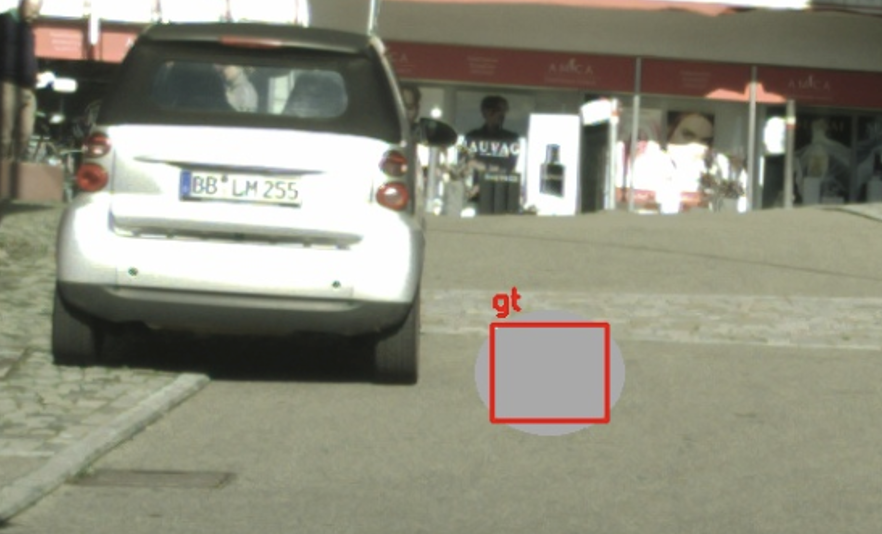}
    \hfill
    \includegraphics[width=0.45\textwidth,height=5.5cm,keepaspectratio]{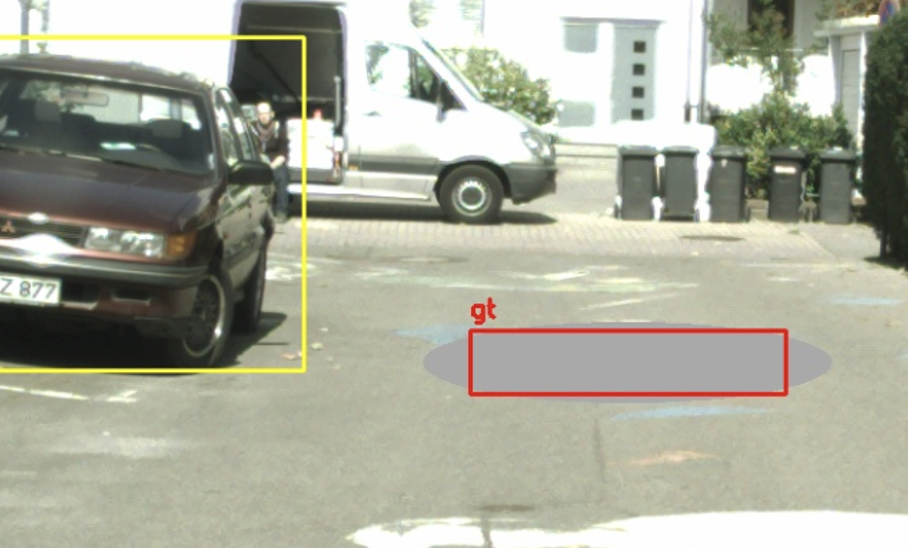}
    \caption*{Grey noise in the form of an oval mask.}

    \vskip\baselineskip

    \includegraphics[width=0.45\textwidth,height=4cm,keepaspectratio]{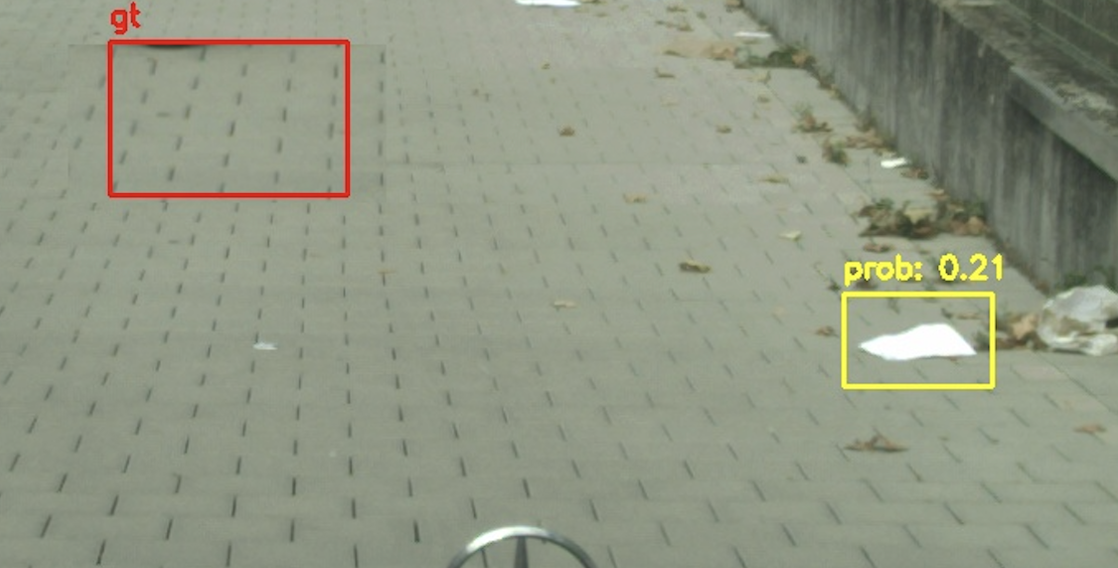}
    \hfill
    \includegraphics[width=0.45\textwidth,height=4cm,keepaspectratio]{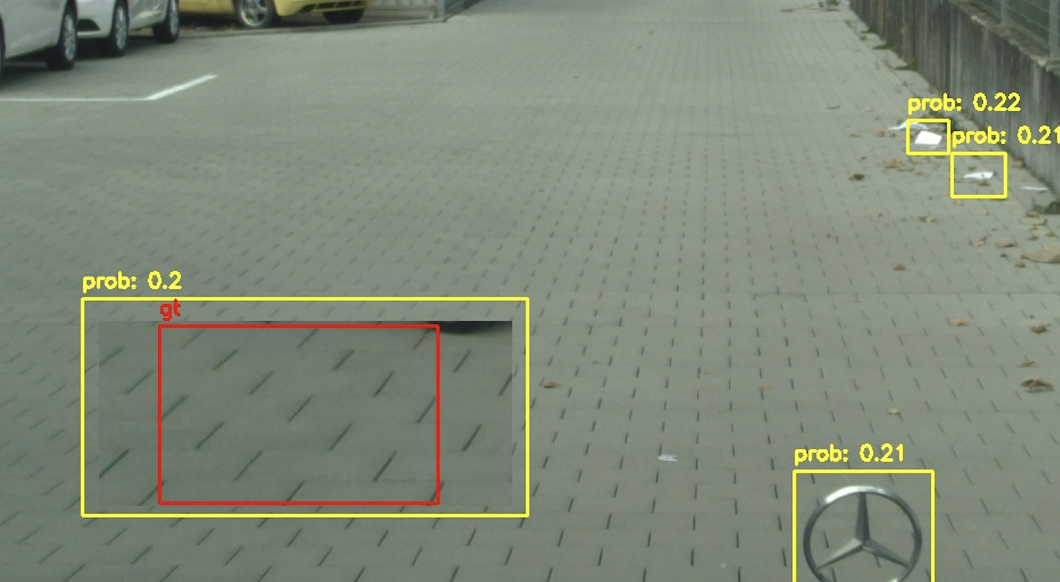}
    \caption*{Copied street pattern from another part of the scene.}

    \vskip\baselineskip

    \includegraphics[width=0.45\textwidth,height=5cm,keepaspectratio]{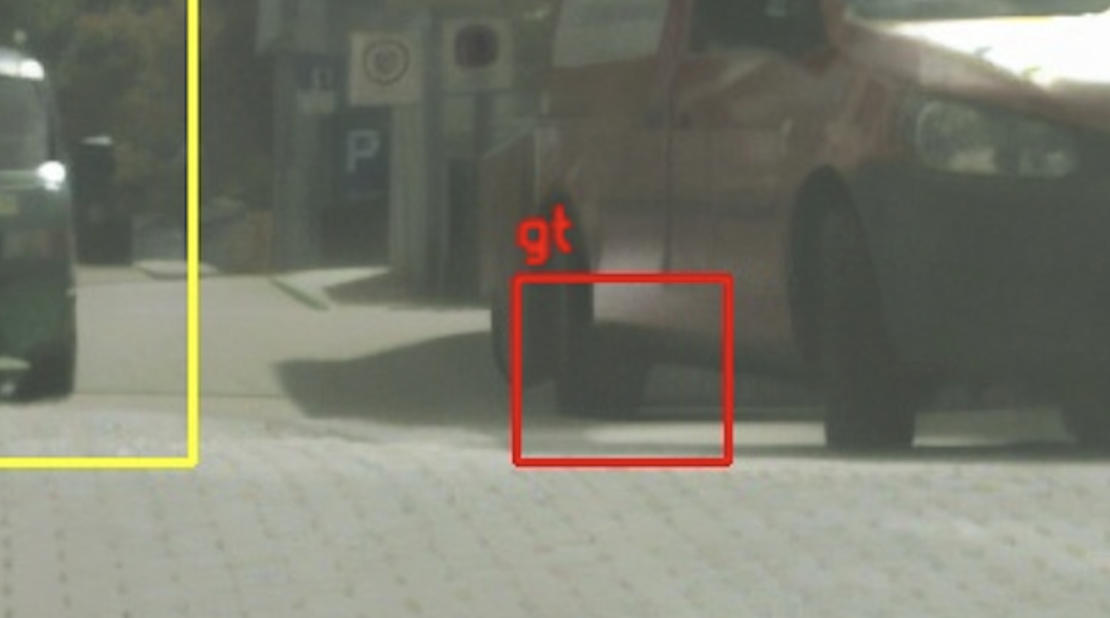}
    \hfill
    \includegraphics[width=0.45\textwidth,height=5cm,keepaspectratio]{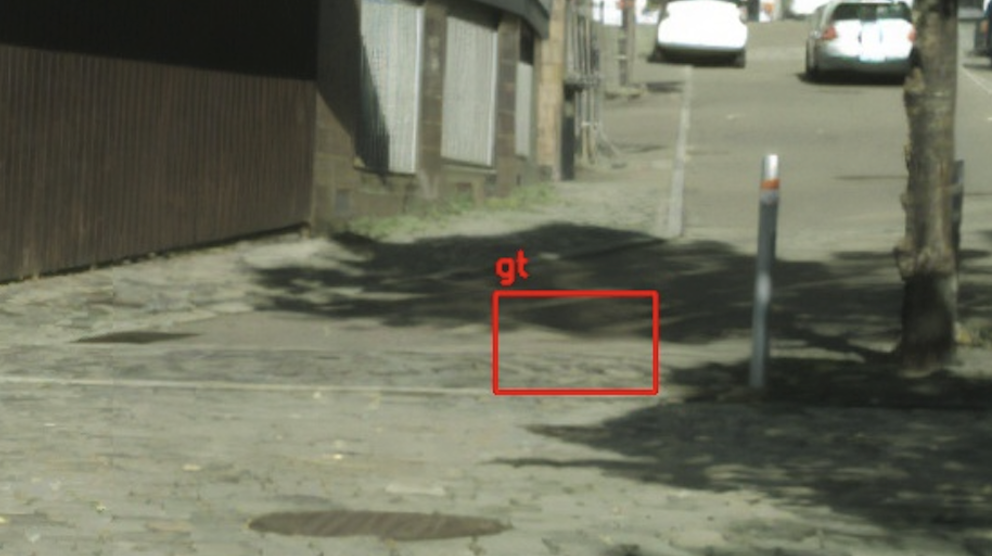}
    \caption*{Removed original object using inpainting.}

    \vskip\baselineskip

    \includegraphics[width=0.45\textwidth,height=4.5cm,keepaspectratio]{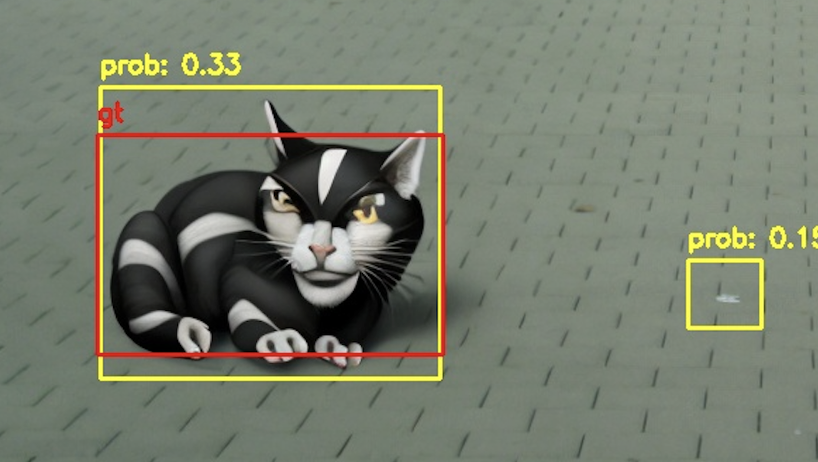}
    \hfill
    \includegraphics[width=0.45\textwidth,height=4.5cm,keepaspectratio]{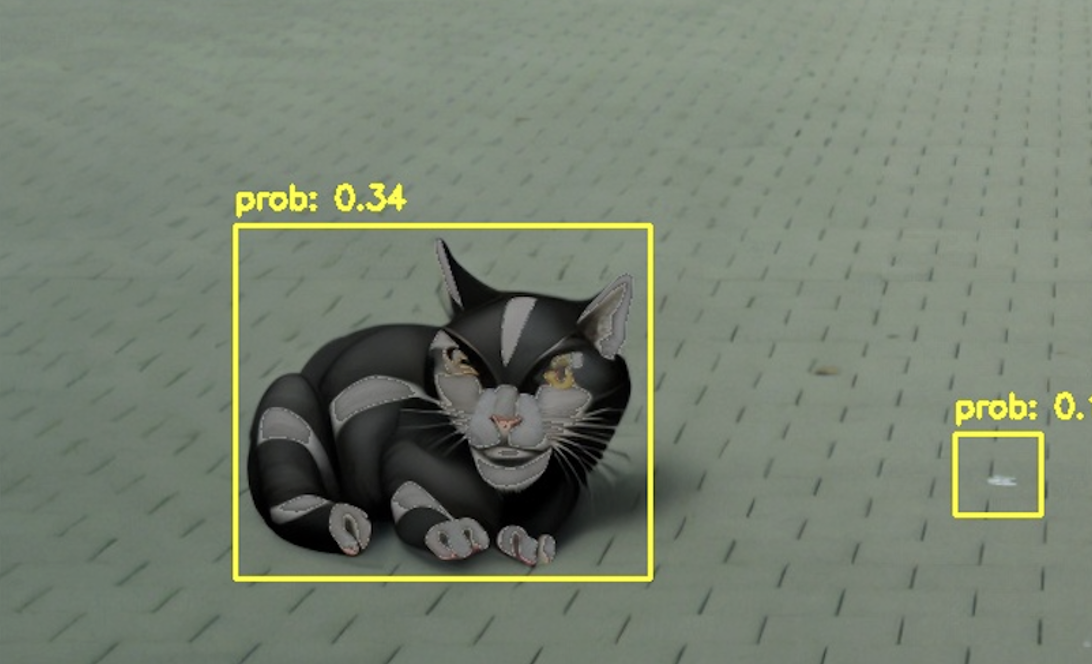}
    \caption*{Smoothed by applying a filter.}

    \caption{Examples of different types of synthetic content added to street scenes. Ground truth bounding boxes are shown in red and predictions in yellow, annotated with the confidence score.}
    \label{fig:additional_examples}
\end{figure}

    \paragraph{Noise.}
    First, we wanted to see how object detectors react to noise in the form of white and grey ovals. For that we placed ovals with the according color based on the location of the ground truth bbox into the streets scene images as shown in a) of \Cref{fig:additional_examples}. This leads to sharp edges between the noise oval and the original content. Since object detectors are trained on common objects, these noise patterns should be hard to detect. However, if correctly predicted anyway, that would indicate a high relevance of the sharp edge for the detection performance of open-vocabulary object detectors, as the noise object shows only one color. 
    
    \paragraph{Pattern.}
    Focused on the sharp edges, we manually placed snippets of the same road pattern onto the ground truth bbox as can be seen in b) of \cref{fig:additional_examples}. Similar to the noise of the previous experiment, again we expect the detector to only detect the `object' in the form of the pattern frame, by recognizing the sharp edges, as the frame itself provides no content. 

    \paragraph{Removed.}
    When visually inspecting the inpaintings of some TP results manually, we could not rarely observe any obvious edges between the original image and the inpainted. However, we wanted to ensure that it is not the inpainted transition between the original image and the inpainted object that leads to a TP detection. In order to prove this, we studied failed inpaintings, i.e.\ those images we filtered out in the main paper which show no actual object, only artificial pixels that are adapted to the surroundings of the masked area.

    \paragraph{Brightness smoothing.}
    After examining some of the inpaintings, we observed that the inpainted content often seemed noticeably brighter than the rest of the image. Therefore, we suspected the different brightness to cause a TP detection independent of the actual object content. To cope with that, we started by calculating the brightness of each pixel in the masked area by following $Brightness = \frac{R+G+B}{3}$ with $R, G, B \in [0,255]$ denoting the three color channels of an RGB image. Afterward, each pixel exceeding the threshold of $Brightness > 200$ has been replaced by the mean color $mC \in [0,255]^{1 \times 3}$ of the surrounding masked area $S \in [0,255]^{n \times m \times 3}$. Herby, $S$ was defined as $S=2M-M$ with $M \in [0,255]^{n \times m \times 3}$ being the original masked area. Accordingly,  $mC = mean(S)$ can be calculated easily. Thus, we retrieved objects integrating more smoothly into the street scene as seen in d) of \cref{fig:additional_examples}.

\section{Findings of additional experiments}
    All additional experiments have been performed with Grounding DINO using the prompt ``object on the street''.
    As it was possible to automatically insert the noise ovals into the street scenes, we did so for the whole LostAndFound Dataset, obtaining 179 images for both white and grey ovals. On the white oval set, the detection led to 175 TP and 4 FN results. For the set with grey ovals, the detector provided 162 TP and 17 FN results. Next, the pattern dataset was tested. As these images were created manually, there are only 16 images. Running a detection on them delivered 1 TP and 15 FN results. We continued with the set showing a synthetic background in the mask area, as the original objects were removed during inpainting. As during the early stage of this work we just had some of these examples, this set contains 21. All of these were detected as FN from Grounding DINO. For the smoothed images we have not conducted quantitative results. However, we saw for particular examples that the confidential score of the prediction was mostly close or even higher compared to the non-smoothed image. One example for this, along with TP and FN results for the other experiments are shown in \cref{fig:additional_examples}.
    Considering the high percentage of TP results on the noise sets brings the assumption that only the edges occur by adding or inpainting synthetic content matters for detection. However, object detectors being sensitive to this kind of edge in images is not surprising. Reflecting on the obvious noise ovals added in this case, we think retrieving some FN indicates that the actual inpainted content itself matters instead of the detector always detecting artificial content as TP. This is supported by the other experiments. Testing on the pattern and removed objects, which denotes synthetic empty content shows as expected and desired almost only FN results across our small experiments. These experiments indicate that inpainting objects is a reasonable way of systematically challenging open vocabulary object detectors.

\section{Additional visualizations}
\begin{figure}[h!]
        \centering

                \begin{minipage}[b]{0.3\linewidth}
                \centering
                \includegraphics[width=\linewidth]{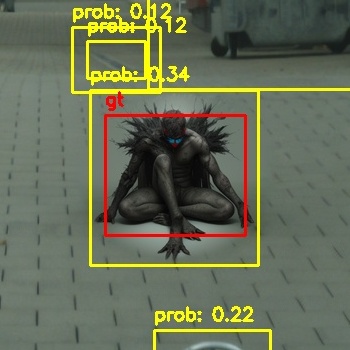}
                \end{minipage}
                \begin{minipage}[b]{0.3\linewidth}
                \centering
                \includegraphics[width=\linewidth]{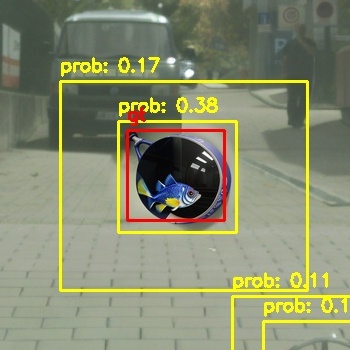}
                \end{minipage}
                \begin{minipage}[b]{0.3\linewidth}
                \centering
                \includegraphics[width=\linewidth]{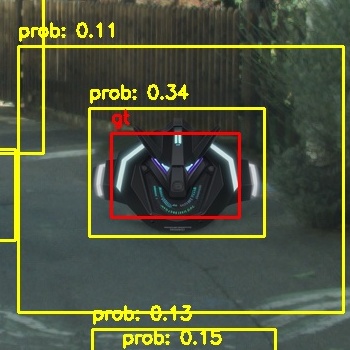}
            \end{minipage}
            
                \begin{minipage}[b]{0.3\linewidth}
                \centering
                \includegraphics[width=\linewidth]{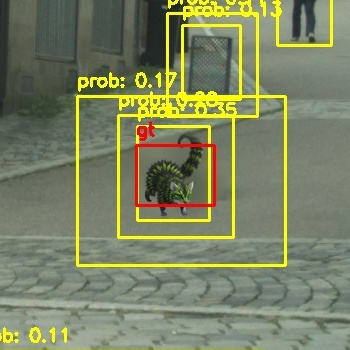}
                \end{minipage}
                \begin{minipage}[b]{0.3\linewidth}
                \centering
                \includegraphics[width=\linewidth]{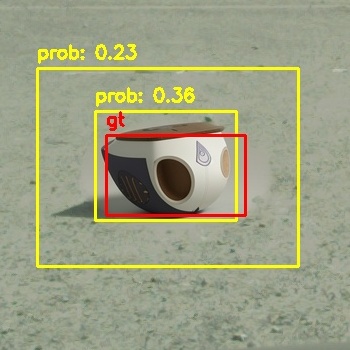}
                \end{minipage}
                \begin{minipage}[b]{0.3\linewidth}
                \centering
                \includegraphics[width=\linewidth]{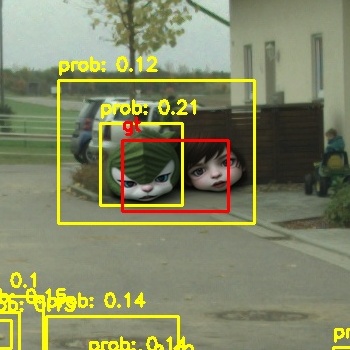}
            \end{minipage}
                \begin{minipage}[b]{0.3\linewidth}
                \centering
                \includegraphics[width=\linewidth]{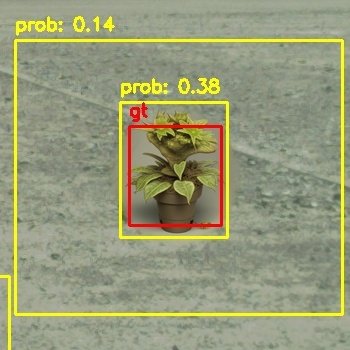}
                \end{minipage}
                \begin{minipage}[b]{0.3\linewidth}
                \centering
                \includegraphics[width=\linewidth]{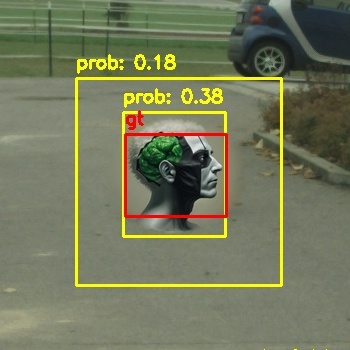}
                \end{minipage}
                \begin{minipage}[b]{0.3\linewidth}
                \centering
                \includegraphics[width=\linewidth]{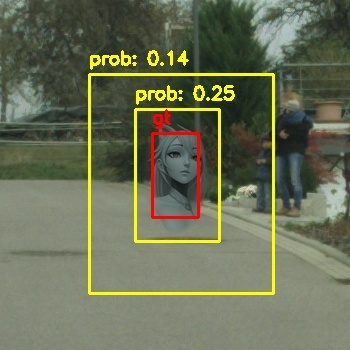}
\end{minipage}
        \caption{Examples for TP predictions using Grounding DINO with prompt ``object on the street'' and a confidence score of 0.1. Red bboxes annotated with gt denoting the ground truth of the original object used for inpainting. Yellow bboxes show model predictions with the respective confidence score.}
        \label{fig:tp_examples}
    \end{figure}

\begin{figure}[htb]
        \centering

                \begin{minipage}[b]{0.3\linewidth}
                \centering
                \includegraphics[width=\linewidth]{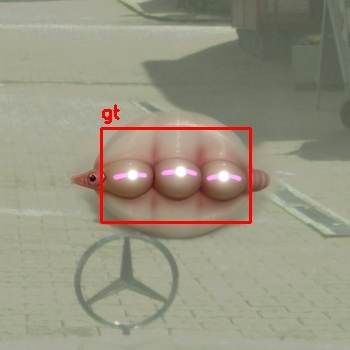}
                \end{minipage}
                                \begin{minipage}[b]{0.3\linewidth}
                \centering
                \includegraphics[width=\linewidth]{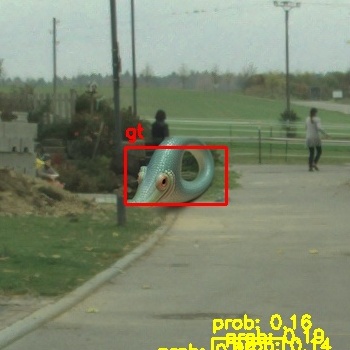}
                \end{minipage}
                                \begin{minipage}[b]{0.3\linewidth}
                \centering
                \includegraphics[width=\linewidth]{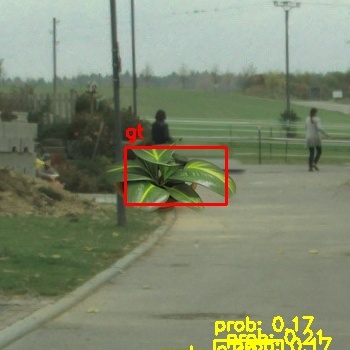}

                \end{minipage}
                                \begin{minipage}[b]{0.3\linewidth}
                \centering
                \includegraphics[width=\linewidth]{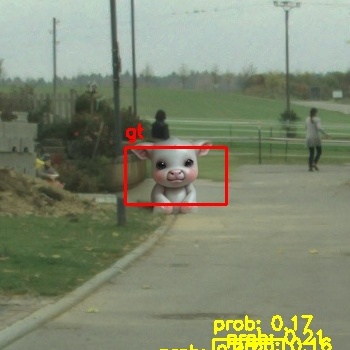}
                \end{minipage}
                                \begin{minipage}[b]{0.3\linewidth}
                \centering
                \includegraphics[width=\linewidth]{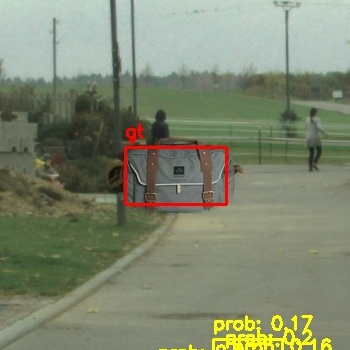}
                \end{minipage}
                                \begin{minipage}[b]{0.3\linewidth}
                \centering
                \includegraphics[width=\linewidth]{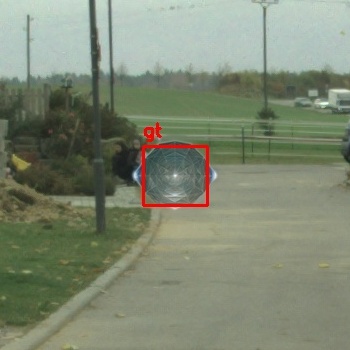}

                \end{minipage}
                                \begin{minipage}[b]{0.3\linewidth}
                \centering
                \includegraphics[width=\linewidth]{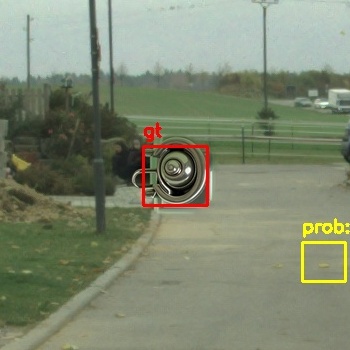}

                \end{minipage}
                                \begin{minipage}[b]{0.3\linewidth}
                \centering
                \includegraphics[width=\linewidth]{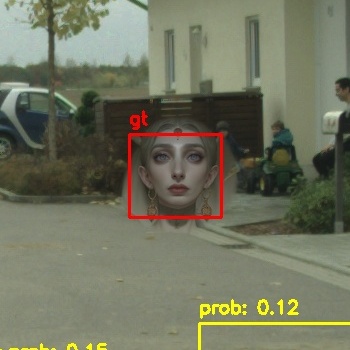}
                \end{minipage}
                                \begin{minipage}[b]{0.3\linewidth}
                \centering
                \includegraphics[width=\linewidth]{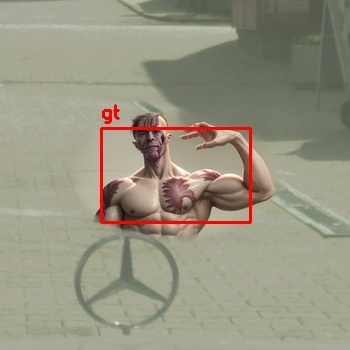}
\end{minipage}
        \caption{Examples for FN predictions using Grounding DINO with prompt ``object on the street'' and a confidence score of 0.1. Red bboxes annotated with gt denoting the ground truth of the original object used for inpainting. Yellow bboxes show model predictions with the respective confidence score.}
        \label{fig:fn_examples}
    \end{figure}

We present random examples of inpainted objects from the generated Hybrid-Concept-Inpainting LostAndFound dataset which were detected correctly by Grounding DINO provided the prompt ``object on the street'' in \cref{fig:tp_examples}. \Cref{fig:fn_examples} depicts objects from the same dataset overlooked by Grounding DINO. Both figures give an idea of the variety and abnormality of the generated data.

\section{Prompt list for Single-Concept-Inpainting}
    prompt\_list = [
    'robot', 'helicopter', 'monster', 'skateboard',
    'dog', 'cat', 'monkey', 'horse', 'elephant', 'lion', 
    'tiger', 'bear', 'deer', 'rabbit', 'squirrel', 'wolf', 
    'fox', 'sheep', 'goat', 'chicken', 'crocodile', 'alligator', 'hamster', 'gerbil', 'mouse', 
    'rat', 'guinea pig', 'ferret', 'rabbit', 'cavy', 'tapir', 
    'hedgehog', 'kangaroo', 'koala', 'panda', 'zebra', 
    'giraffe', 'hippopotamus', 'rhinoceros', 'sloth', 'antelope', 
    'bison', 'buffalo', 'ostrich', 'emu','penguin', 'seal', 'walrus', 'manatee', 'platypus', 'okapi', 
    'armadillo', 'badger', 'mole', 'opossum', 'raccoon', 
    'porcupine', 'weasel', 'lemur', 'gorilla', 'chimpanzee', 
    'orangutan', 'tamarin', 'sloth bear', 'sea lion', 'tortoise',
    'flamingo', 'robot', 'helicopter', 'monster', 'skateboard',
    "Sofa", "Coffee table",  "Bookshelf", "Lamps",
    "Cutting board", "Pots pans", "Dishes",
    "Desk", "Chair", "Printer",
    "Vacuum cleaner", "Fan", "Clock", "Shoes"
    ]

\end{document}